\documentclass[11pt,final]{article}

\pdfoutput=1

\usepackage[final]{acl}

\usepackage{times}
\usepackage{latexsym}
\usepackage{graphicx}
\usepackage{multirow}
\usepackage{booktabs} 
\usepackage{hyperref}
\usepackage{tabularx}
\usepackage{float} 
\usepackage[T1]{fontenc}
\usepackage[utf8]{inputenc}
\usepackage{microtype}
\usepackage{inconsolata}
\usepackage{amsmath}
\usepackage{amssymb}
\usepackage{adjustbox}
\usepackage{multirow, makecell}
\usepackage{booktabs} 

\title{HICD: Hallucination-Inducing via Attention Dispersion for Contrastive Decoding to Mitigate Hallucinations in Large Language Models}

\author{
Xinyan Jiang\textsuperscript{1,2}, Hang Ye\textsuperscript{1,2}, Yongxin Zhu\textsuperscript{1}\thanks{Corresponding author}, Xiaoying Zheng\textsuperscript{1}, Zikang Chen\textsuperscript{1,2}, Jun Gong\textsuperscript{1,2} \\
\textsuperscript{1}Shanghai Advanced Research Institute, Chinese Academy of Sciences, Shanghai, China \\
\textsuperscript{2}University of Chinese Academy of Sciences, Beijing, China \\
\texttt{\{jiangxy2024, zhuyongxin\}@sari.ac.cn}
}

\begin{document}
\maketitle
\begin{abstract}
Large Language Models (LLMs) often generate hallucinations, producing outputs that are contextually inaccurate or factually incorrect. We introduce \textbf{HICD}, a novel method designed to induce hallucinations for contrastive decoding to mitigate hallucinations. Unlike existing contrastive decoding methods, HICD selects attention heads crucial to the model's prediction as inducing heads, then induces hallucinations by dispersing attention of these inducing heads and compares the hallucinated outputs with the original outputs to obtain the final result. Our approach significantly improves performance on tasks requiring contextual faithfulness, such as context completion, reading comprehension, and question answering. It also improves factuality in tasks requiring accurate knowledge recall. We demonstrate that our inducing heads selection and attention dispersion method leads to more "contrast-effective" hallucinations for contrastive decoding, outperforming other hallucination-inducing methods. Our findings provide a promising strategy for reducing hallucinations by inducing hallucinations in a controlled manner, enhancing the performance of LLMs in a wide range of tasks.\footnote{ \url{https://github.com/waitxian/HICD.git}}

\end{abstract}

\section{Introduction}

Large language models(LLMs) have demonstrated exceptional performance across a wide range of NLP tasks \cite{NEURIPS2020_1457c0d6,wang-etal-2024-factuality}. However, they are prone to hallucinations, where they generate content that deviates from facts or relevant contexts, hindering their practical applications in real-world scenarios. To address this challenge, efforts have been devoted to mitigate knowledge hallucinations in LLMs \cite{NEURIPS2022_8bb0d291,dhuliawala2023chain}. In this work, we focus on mitigating hallucinations during inference generation \cite{li-etal-2024-dawn}.

To address this, some studies have focused on developing effective inference-time decoding strategies. Among these, contrastive decoding based approaches have demonstrated strong performance \cite{shi-etal-2024-trusting}. However, current contrastive decoding methods typically compare the model's inherent outputs, such as those from earlier layers or smaller models, with the original outputs\cite{chuang2024dola,li-etal-2023-contrastive}. Existing contrastive decoding approaches have rarely explored constructing hallucinated outputs to improve their efficacy in hallucination mitigation\cite{sahoo-etal-2024-comprehensive}.

Previous work has highlighted that current contrastive decoding methods, due to their coarse contrast and simplistic subtraction operations, may disrupt the original output distribution of the LLM \cite{chen2024lower}. Therefore, investigating the construction of hallucinated outputs for more effective contrast with original outputs warrants further research. Building on this, \cite{Zhang2023AlleviatingHO} proposed inducing hallucinations in LLMs via slight fine-tuning or zero-shot prompting, and mitigating them through contrastive decoding with the original outputs. And there's a method that prunes retrieval heads to generate hallucinated outputs for comparison with the original outputs \cite{gema2024decore}. However, these hallucination-inducing methods require additional fine-tuning or rely on the inherent properties of the model post-pretraining, limiting their adaptability in different datasets. Moreover, the plausibility of the hallucinations and their effectiveness for contrastive decoding have not been validated.
\begin{figure*}[t] 
    \centering
    \includegraphics[width=\textwidth]{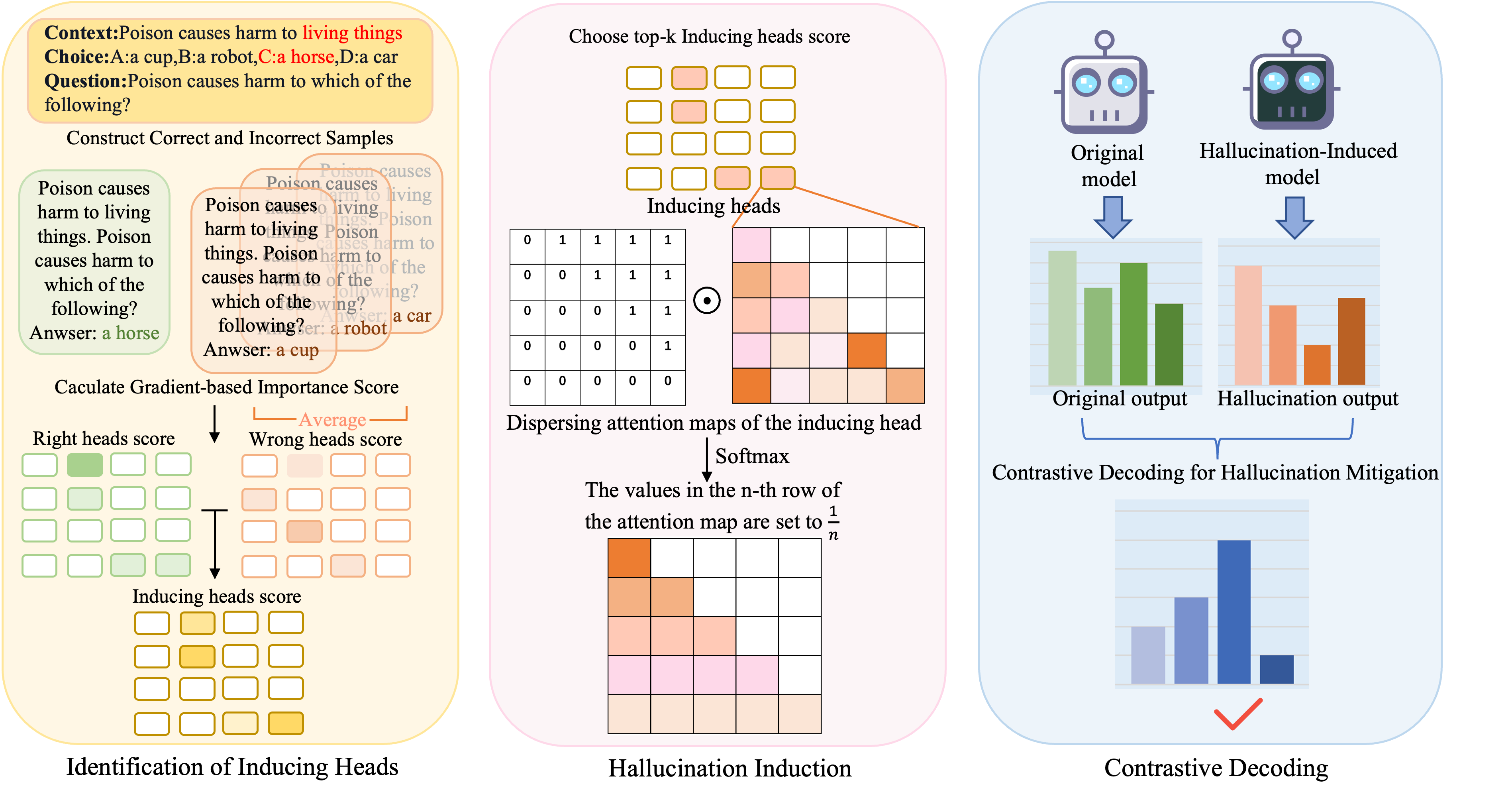} 
    \caption{Illustration of \textbf{H}allucination-\textbf{I}nducing \textbf{C}ontrastive \textbf{D}ecoding Method\textbf{(HICD)}. The method include calculation of the importance scores and identification of the inducing heads (yellow), dispersing attention of inducing heads to induce hallucinations (pink) and applying contrastive decoding for hallucination mitigation (blue).}

    \label{fig:example}
\end{figure*}

Other works have addressed the issue of hallucinations by focusing on model interpretability. Some studies examined attention heads that play a key role in output quality \cite{bansal-etal-2023-rethinking}. Another study revealed that key points causing hallucinations in LLMs are the inconsistencies in the information flow integration between memory heads and context heads, and effectively mitigated hallucinations by pruning conflicting attention heads \cite{jin-etal-2024-cutting}. This suggests that targeting the attention heads critical to hallucinated outputs can effectively control hallucination generation. 

Inspired by these studies,we propose \textbf{HICD}, \textbf{\textit{a method that induces hallucinations through attention dispersion on inducing heads for contrastive decoding to mitigate hallucinations}}. To address the limitation that existing hallucination-inducing methods rely on model's internal parameters, restricting adaptability to different datasets, we construct correct and incorrect (adversarial) samples by pairing questions with corresponding right and wrong answers. We then compute task-relevant importance scores for attention heads that are critical to generating correct outputs (\textit{right heads}) and incorrect outputs (\textit{wrong heads}). Finally, we select heads that contribute to correct outputs while suppressing those leading to incorrect outputs, resulting in a set of \textit{inducing heads}.

To improve the effectiveness of contrastive decoding methods, the attention maps of the inducing heads are averaged, ensuring attention values are equalized across all tokens within each head. This redistribution disperses attention, effectively inducing hallucinated outputs optimized for contrastive decoding, as demonstrated by experiments. Finally, these hallucinated outputs are compared with the original model's outputs to mitigate hallucinations.

Our experiments are primarily conducted using models from the LLaMA~\cite{touvron2023bllama}, Qwen~\cite{bai2023qwen}, and Mistral~\cite{jiang2023mistral7b} families. Our findings show that compared to existing contrastive decoding methods, HICD significantly improves faithfulness in tasks requiring contextual understanding, such as HellaSwag\cite{zellers2019hellaswag}, RACE\cite{lai-etal-2017-race}, OpenBookQA \cite{mihaylov-etal-2018-suit}. Furthermore, HICD also enhances the model’s accuracy in factual recall tasks like TruthfulQA\cite{lin-etal-2022-truthfulqa} and Factor\cite{muhlgay-etal-2024-generating}, as well as generation tasks on XSum \cite{chuang-etal-2024-lookback} and NQ-Swap \cite{longpre-etal-2021-entity}. Our contributions are as follows:
\begin{itemize}
    \item \textbf{Task-Driven Inducing Head Selection:} Inducing heads selected based on task, yield more effective hallucination induction than task-irrelevant selecting methods.
    \item\textbf{Attention Dispersion:} Averaging the attention maps of inducing heads increases the effectiveness of hallucinated outputs by allowing context with lower relevance to the prediction to influence the results.
    \item \textbf{Contrast Effective:} HICD leads to more effective hallucination outputs and better mitigation during contrastive decoding.
\end{itemize}

\section{Background}

\subsection{Multi-head Attention}
Multi-head attention is crucial in transformer-based models, enabling them to capture complex dependencies by attending to different parts of the input sequence simultaneously
\cite{halawi2023overthinking}. 

Formally, given the input sequence \( x^{\ell-1} = [x_1^{\ell-1}, \ldots, x_N^{\ell-1}] \) at layer \( \ell \), an MHA block in the transformer computes a set of attention heads. Each attention head \( h \) at layer \( \ell \) is computed as follows:

\begin{equation}
s^{\ell,h} = \sigma \left( \frac{(X^{\ell-1} W_Q^{\ell,h}) (X^{\ell-1} W_K^{\ell,h})^T}{\sqrt{d/M}} \right) \label{eq:mha_head}
\end{equation}

where \( X^{\ell-1} \in \mathbb{R}^{N \times d} \) represents the input hidden states, \( d \) is the dimensionality, and \( M \) is the number of heads. \( W_Q^{\ell,h} \), \( W_K^{\ell,h} \), and \( W_V^{\ell,h} \) are the queries, keys, and values for the \( h \)-th head, respectively. The attention score is the dot product of queries and keys, scaled by \( \sqrt{d/M} \), and passed through the softmax function \( \sigma \) to get the attention distribution \( s^{\ell,h} \).

The final attention output for the \( h \)-th head, \( H^{\ell,h} \), is computed by:

\begin{equation}
H^{\ell,h} = s^{\ell,h} X^{\ell-1} W_V^{\ell,h}
\label{eq:H}
\end{equation}

The attention output of all heads is then concatenated to form the output of the MHA block:

\begin{equation}
A^{\ell} = [H^{\ell,1}; H^{\ell,2}; \ldots; H^{\ell,M}] W_O^{\ell} \label{eq:mha_output}
\end{equation}

where \( W_O^{\ell} \) is a learnable output matrix that projects the concatenated attention heads back to the desired dimensionality.

\subsection{Gradient-based Importance Score}

The gradient-based importance score quantifies the contribution of an attention head \( h \) to the model's predictions by calculating the sensitivity of the output to changes in \( h \) \cite{NEURIPS2019_2c601ad9,bansal-etal-2023-rethinking}. Given a dataset \( \mathcal{D} \), the score is computed as:
\begin{equation}
    I_h(\mathcal{D}) = \mathbb{E}_{(x,y) \sim \mathcal{D}} \left| \frac{\partial \mathcal{L}(y|x)}{\partial A^h([x; y])} \right| \label{eq:gradient_score}
\end{equation}

where \( \mathcal{L}(y|x) \) is the loss function, \( A^h([x; y]) \) is the output of attention head \( h \), and \( (x, y) \) are input-output pairs from \( \mathcal{D} \). The model’s loss is computed using the negative log-likelihood:
\begin{equation}
    \mathcal{L}(y|x) = -\frac{1}{T_y} \sum_{j=1}^{T_y} \log p(y_j | x, y_{1:j-1}) 
    \label{eq:log_likelihood}
\end{equation}

The importance scores for all heads are efficiently computed by performing a single forward and backward pass over the model with \( \mathcal{D} \).

\section{Method}
The overall algorithm of HICD is shown in Figure \ref{fig:example}. First, we identify the inducing heads that are closely associated with generating hallucinations(\ref{sec:identification}). Next, we apply attention dispersion to these inducing heads to induce task-relevant hallucinations (\ref{sec:redistribution}). Finally, these hallucinated outputs are compared with the original model outputs through contrastive decoding to alleviate hallucinations (\ref{sec:contrastive_decoding}).

\subsection{Identification of Inducing Heads}\label{sec:identification}
To discover the attention heads that are crucial for correct and incorrect outputs on different datasets, we define a process for identifying the final set of inducing heads. We begin by constructing an adversarial dataset \( D_m' \) based on the original dataset \( (x, c) \in T_m \), where 
\( m \) refers to the specific task, \( x \) represents the context, \( c \) denotes a set of answer choices. Given a dataset \( (x, c, y_i) \in D_m \), where \( y_i \) is the right anwser that belongs to one of the choices \( c \), and we generate the new sample \( (x, c, y_j) \in D_m' \), where \( y_j \in c\ \setminus \{y_i\} \). This results in adversarial samples that pair questions with incorrect answers, derived from the original dataset \( T_m \).

Utilizing both the correct and adversarially constructed incorrect samples, we compute the gradient-based importance score for each attention heads, as defined in Equation \ref{eq:gradient_score}. Based on these importance scores \( I_{l,h}(D_m) \) and \( I_{l,h}(D_m') \), we define a discrepancy correction factor \( F_{l,h}^m \) as:
\begin{equation}
\hspace{-2mm} F_{l,h}^m = I_{l,h}(D_m) - \frac{1}{|c\ \setminus \{y_i\}|} \sum_{y_j} I_{l,h}(D_m') 
\end{equation}

where \( I_{l,h}(D_m) \) and \( I_{l,h}(D_m') \) represent the importance scores in \( D_m \) and \( D_m' \), respectively, with \( l \) referring to the layer and \( h \) representing the attention head. The term \( |c \setminus \{y_i\}| \) represents the size of the set \( c \) excluding the correct answer \( y_i \). The final inducing heads score in dataset \( T_m \) is defined as:
\begin{equation}
S_{l,h}^m(D_m, D_m') = I_{l,h}(D_m) - s \cdot F_{l,h}^m
\end{equation}

\noindent where \( s \) is a hyperparameter scaling factor that controls the influence of the discrepancy between right and wrong heads on the inducing heads score. We select the top \( k_m \) attention heads based on the inducing heads score from dataset \( T_m \). The optimal number \( k_m \) of inducing heads for each dataset is determined experimentally, as described in \ref{sec:experiments}. More details are shown in Appendix \ref{sec:Experimental Setup Details}.

\subsection{Attention Dispersion for Hallucination Induction}\label{sec:redistribution}
We perform attention map averaging on the inducing heads obtained in Section~\ref{sec:identification}. Specifically, given the query \( Q^{\ell,h} \) and key \( K^{\ell,h} \) of an inducing head \( h \) at layer \( \ell \), we apply a lower triangular mask \( M^{\ell,h} \) such that:
\begin{equation}
M^{\ell,h}_{ij} = 
\begin{cases} 
0 & \text{if } i \geq j, \\
1 & \text{if } i < j
\end{cases}
\end{equation}
This mask is multiplied element-wise with the product of \( Q^{\ell,h} \) and \( K^{\ell,h} \) to generate a modified query-key interaction matrix based on Equation \ref{eq:mha_head}:
\begin{equation}
\alpha_{\text{new}}^{\ell,h} = M^{\ell,h} \odot \frac{(Q^{\ell,h} (K^{\ell,h})^T)}{\sqrt{d/M}}
\end{equation}
where \( \odot \) represents the element-wise multiplication operation. This operation forces the lower triangular part of \( \alpha_{\text{new}}^{\ell,h} \) to become zero. Then, in Equation \ref{eq:softmax}, applying the softmax operation \( \sigma \), the attention values for each position are equalized, with all entries in the lower triangular part of the attention map being set to \( \frac{1}{n} \), where \( n \) refers to the index of the row in the attention matrix:
\begin{equation}
s_{\text{inducing}}^{\ell,h} = \text{\( \sigma \)}(\alpha_{\text{new}}^{\ell,h})
\label{eq:softmax}
\end{equation}
\begin{equation}
H_{\text{inducing}}^{\ell,h} = s_{\text{inducing}}^{\ell,h} X^{\ell-1} W_V^{\ell,h}
\label{eq:induced}
\end{equation}
Then, \( s_{\text{inducing}}^{\ell,h} \) is substituted into Equation \ref{eq:H} to get Equation \ref{eq:induced}. After obtaining \( H_{\text{inducing}}^{\ell,h} \) , the model's attention towards each token position in the inducing head is equalized, thus achieving attention dispersion, with the processed model called \textit{induced model}. Experiments in \ref{sec:more_analysis} demonstrate dispersing attention in inducing heads induces more effective hallucination outputs for contrastive decoding.

\subsection{Contrastive Decoding for Hallucination Mitigation}\label{sec:contrastive_decoding}

Given the induced model from Section~\ref{sec:redistribution}, the goal of this approach is to mitigate hallucination in the generated output. We propose a contrastive decoding approach that contrasts the token distributions from the base model and the induced model,which is defined as a re-weighting of the next-token distributions of the base model and the induced model.
\begin{equation}
p(x_t | x_{<t}) \propto \exp \left[ (1 + \alpha) \log p_{\text{original}}(x_t | x_{<t}) \right.
\label{eq:contrastive_decoding}
\end{equation}
\[
\left. - \alpha \log p_{\text{inducing}}(x_t | x_{<t}) \right]
\]

In Equation~\ref{eq:contrastive_decoding}, the new next-token distribution \( p(x_t | x<t) \) is derived by contrasting the next-token distributions of the original model \( p_{\text{original}}(x_t | x<t) \) and the induced model \( p_{\text{inducing}}(x_t | x<t) \).

The scaling factor \( \alpha \in \mathbb{R} \) controls the relative influence between the original and induced models. When \( \alpha > 0 \), the likelihood of the original model is emphasized, leading to a preference for token predictions consistent with the output of the original model. And the likelihood of the induced model is penalized by the term \( \alpha \log p_{\text{inducing}}(x_t | x<t) \), which discourages the selection of tokens that are likely under the induced model.

\begin{table*}[t]
\centering
\small
\caption{Performance of different models and methods on faithfulness evaluation tasks. The best performance is indicated in \textbf{bold}, and the second-best is \underline{underlined}. "*" means we report results of previous research. The results on other models can be found in Appendix \ref{sec: G appendix}. The hyperparameter settings are provided in Table \ref{tab:final_hyperparams}.}
\label{table1}
\begin{tabular}{lcccccccccc}
\toprule
\multirow{2}{*}{\textbf{Backbone}} & \multirow{2}{*}{\textbf{Methods}} & \textbf{Hellaswag} & \multicolumn{2}{c}{\textbf{Race}} & \multicolumn{2}{c}{\textbf{HaluEval-Sum}} & \textbf{OpenbookQA} \\
\cmidrule(lr){3-3} \cmidrule(lr){4-5} \cmidrule(lr){6-7} \cmidrule(lr){8-8}
& & \textbf{Acc} & \textbf{Middle} & \textbf{High} & \textbf{Acc\_H} & \textbf{Acc\_A} & \textbf{Acc} \\
\midrule
\multirow{5}{*}{LLaMA-7b} & Vanilla & 0.7761 & 0.5642 & 0.4339 & 18.94 & 26.06 & 0.5142 \\
& +Alpaca & \underline{0.7849} & \underline{0.5947} & \textbf{0.4806} & 18.31* & \textbf{37.24}* & 0.4901 \\
& +DoLa & 0.7517 & 0.5710 & 0.4462 & \underline{20.41} & 25.91 & 0.4845 \\
& +CAD & - & 0.5772 & 0.4522 & - & - & \underline{0.5463} \\
& +HICD (Ours) & \textbf{0.8423} & \textbf{0.5989} & \underline{0.4668} & \textbf{27.15} & \underline{27.25} & \textbf{0.5581} \\
\midrule
\multirow{4}{*}{LLaMA2-7b} & Vanilla & \underline{0.7832} & 0.5801 & 0.43253 & 24.27 & 48.9 & 0.4846 \\
& +DoLa & 0.6925 & 0.5536 & 0.4070 & 27.78 & 50.31 & 0.4941 \\
& +CAD & - & \underline{0.5898} & \textbf{0.4545} & - & - & \textbf{0.5302} \\
& +HICD (Ours) & \textbf{0.8433} & \textbf{0.5996} & \underline{0.4514} & \textbf{37.46} & \textbf{52.65} & \underline{0.5223} \\
\bottomrule
\end{tabular}
\vspace{-0.1cm}
\end{table*}

\section{Experiments}
\subsection{Experimental Setup}

\textbf{Datasets and Metrics.} \textbf{1) Faithfulness evaluation:} For context completion, we evaluate on HellaSwag\cite{zellers2019hellaswag}, where the goal is to predict the next sentence based on context. For reading comprehension (RACE-H and RACE-M\cite{lai-etal-2017-race}), representing high school and middle school levels. For question answering, we use the additional subset of OpenBookQA\cite{mihaylov-etal-2018-suit} with a "fact1" field as reference context. \textbf{2) Knowledge hallucination:} Evaluated with HaluEval-Sum\cite{li-etal-2023-halueval}, using accuracy for both hallucinated and correct summaries (Acc-A and Acc-H). \textbf{3) Factuality evaluation:} Done with TruthfulQA\cite{lin-etal-2022-truthfulqa} and Factor\cite{muhlgay-etal-2024-generating}, measuring the model's ability to provide truthful answers (TruthfulQA) and generate factual completions (Factor). \textbf{4) Open-ended generation:} We use XSum~\cite{xsum-emnlp} for evaluating summarization quality, and NQ-Swap~\cite{longpre-etal-2021-entity} for assessing contextual faithfulness in open-ended generation.
\vspace{1em}

\noindent\textbf{Models and Baselines.} Our experiments are mainly conducted using Llama models. 
We compare HICD with the following decoding methods: \textbf{1) greedy decoding}, which
greedily selects the next token with the highest probability; \textbf{2) DoLa}\cite{chuang2024dola}, which attempts to reduce hallucinations by contrasting output distributions from different layers of the model; \textbf{3) Contrastive decoding (CD)}\cite{li-etal-2023-contrastive}, which contrasts
output distributions from models of different scales of parameters; \textbf{4) Context-Aware Decoding (CAD)}\cite{shi-etal-2024-trusting}, a variant of CD where the amateur model is the same as the expert model but is not presented with the additional context.
Details of experimental setups and datasets are provided in Appendix \ref{sec: A appendix}.

\subsection{Main Results}

\textbf{HICD Mitigates Faithfulness Hallucinations.} Table~\ref{table1} presents the performance of different contrastive decoding methods in faithfulness-related tasks. HICD outperforms other methods in all tasks, showing significantly better contextual faithfulness. It achieves the highest or second-highest scores across tasks, with additional results in other models provided in Appendix~\ref{sec: G appendix}. Detailed parameter settings are provided in Appendix \ref{sec: B appendix}.

For example, HICD achieves 84.23\% accuracy on the HellaSwag context completion task with Llama-7B, a 6.6\% improvement over greedy decoding and a significant improvement compared to other methods. It also performs well on reading comprehension and question answering tasks, surpassing other methods on the RACE benchmark and achieving competitive results on OpenBookQA. In the HaluEval-Sum knowledge hallucination task, HICD achieves significant improvements with Llama2-7B, scoring 37.46 (Acc-H) and 52.65 (Acc-A), outperforming the next best results by 9.7\% and 2.3\%, respectively. Additionally, with Llama2-7B, HICD outperforms CAD on RACE-Middle, and scores comparably to CAD on RACE-High and OpenBookQA, securing the second-best performance.\vspace{1em}

\begin{figure*}[t]
\centering
\includegraphics[width=\textwidth]{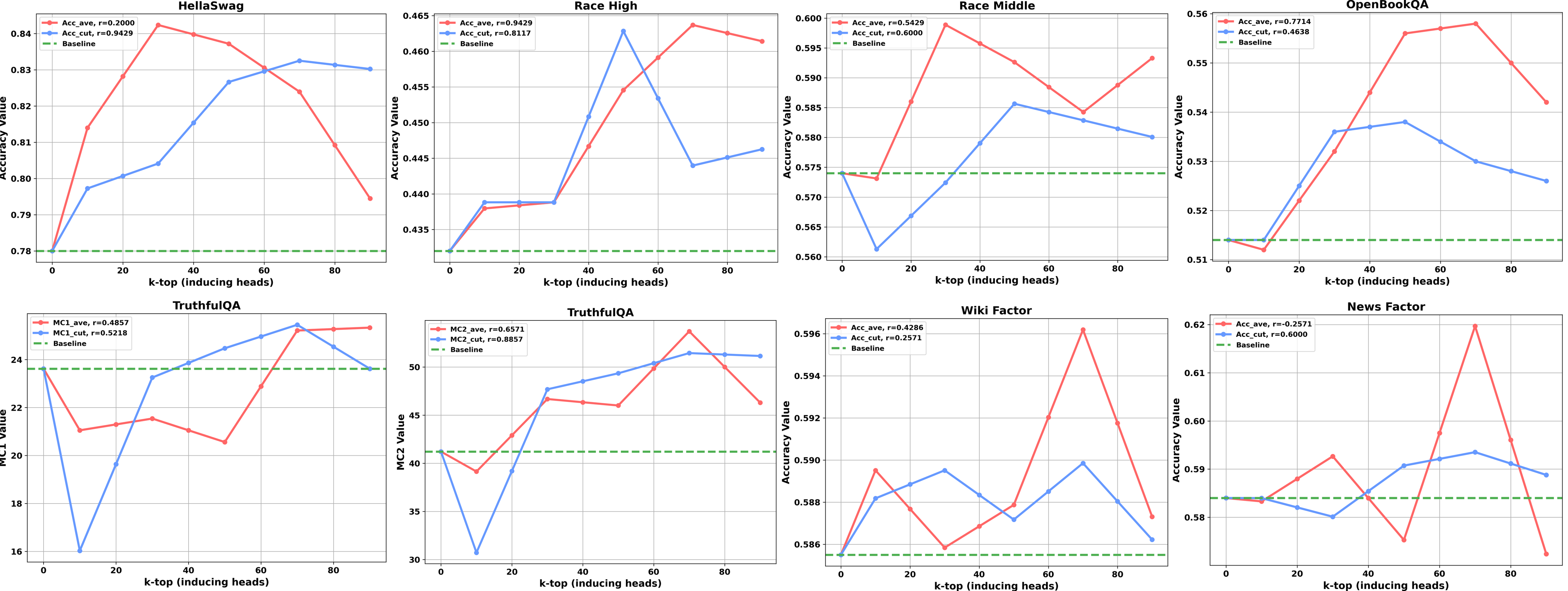} 
\caption{
Effect of inducing head number on task performance. The red lines represent our HICD method, using average attention over inducing heads to induce hallucinations. The blue lines show the head-pruning method from prior research, where inducing heads are pruned (implementation details in Appendix \ref{sec:head_pruning}). The green dashed line represents the baseline model without hallucination induction. Spearman correlation coefficient $r$ measures the correlation between inducing heads and task performance. The parameter \( \alpha  \) and\( \ s \) tuning are shown in Appendix ~\ref{sec: B appendix}.
}
\label{fig:induced_heads_analysis}
\vspace{-0.5cm}
\end{figure*}

\begin{table}[t]
\centering
\small
\caption{Performance of different decoding methods on factuality evaluation tasks. The best performance is indicated in \textbf{bold}, the second-best is \underline{underlined}. "*" means we report results of previous research.}
\label{table:performance}
\begin{tabular}{l@{\hskip 5pt}p{0.8cm}@{\hskip 5pt}p{0.8cm}@{\hskip 5pt}p{0.8cm}@{\hskip 8pt}p{1.2cm}@{\hskip 5pt}p{1.2cm}}
\toprule
\multirow{2}{*}{\textbf{Methods}} & \multicolumn{3}{c}{\textbf{TruthfulQA}} & \multicolumn{2}{c}{\textbf{FACTOR}} \\
\cmidrule(lr){2-4} \cmidrule(lr){5-6}
& \textbf{MC1} & \textbf{MC2} & \textbf{MC3} & \multicolumn{1}{c}{\textbf{WIKI}} & \multicolumn{1}{c}{\textbf{NEWS}} \\
\midrule
LLaMA-7b & 23.62 & 41.21 & 19.33 & \multicolumn{1}{c}{0.5855} & \multicolumn{1}{c}{0.5840} \\
+Alpaca & 22.88 & \underline{52.47} & 25.19 & \multicolumn{1}{c}{0.5711} & \multicolumn{1}{c}{0.5820} \\
+DoLa & \textbf{31.95} & 52.21 & \textbf{28.17} & \multicolumn{1}{c}{\underline{0.6196}} & \multicolumn{1}{c}{0.6168} \\
+13b-CD & 24.40 & 41.01 & 19.03 & \multicolumn{1}{c}{\textbf{0.6411}} & \multicolumn{1}{c}{\underline{0.6190}} \\
+HICD  & \underline{25.45} & \textbf{53.71} & \underline{26.52} & \multicolumn{1}{c}{0.6058} & \multicolumn{1}{c}{\textbf{0.6197}} \\
\midrule
LLaMA2-7b & \underline{28.51} & 43.30 & 22.40 & \multicolumn{1}{c}{0.5898} & \multicolumn{1}{c}{0.7203} \\
+DoLa & \textbf{34.51} & \textbf{55.91} & \underline{28.81} & \multicolumn{1}{c}{\textbf{0.6325}} & \multicolumn{1}{c}{\underline{0.7268}} \\
+13b-CD & 28.15* & \underline{54.87}* & \textbf{29.75}* & \multicolumn{1}{c}{-} & \multicolumn{1}{c}{-} \\
+HICD  & 23.99 & 51.28 & 25.89 & \multicolumn{1}{c}{\underline{0.6069}} & \multicolumn{1}{c}{\textbf{0.7346}} \\
\bottomrule
\end{tabular}
\vspace{-0.2cm}
\end{table}

\noindent\textbf{HICD Mitigates Factuality Hallucinations.} Although HICD’s primary goal is to improve contextual faithfulness by mitigating hallucinations, its effectiveness in factual consistency tasks remains an open question. Therefore, we also evaluate HICD on TruthfulQA and Factor tasks, where the model is required to generate factually accurate outputs. Besides comparing with the previously mentioned baselines, we also compare with the model fine-tuned on the Alpaca dataset\cite{alpaca}.

In Table~\ref{table:performance}, we can see that HICD improves the accuracy of the model in factual consistency tasks. Specifically, on the multiple choice task in TruthfulQA, with Llama-7B, HICD achieves competitive results across all metrics compared to the baselines, surpassing DoLa and Alpaca on the MC2 metric. In the Factor task, for all models, although HICD achieves slightly lower scores compared to 13B-CD and DoLa in Wiki dataset, it achieves the highest score in the News Factor dataset. More detailed results analyses are shown in Appendix \ref{sec: C appendix}.
\label{sec:experiments}

\noindent\textbf{HICD on open-ended generation scenarios.} Following the experimental protocol in~\cite{chuang-etal-2024-lookback}, we constructed a test set by sampling 1,000 examples from the XSum\cite{xsum-emnlp} dataset. Since the dataset does not include hallucinated summaries, we generated them using GPT-4 to construct contrastive instances for inducing head selection. We also used the NQ-Swap \cite{longpre-etal-2021-entity}, which provides entity-swapped contexts, allowing evaluation of contextual factuality.

As results in Table~\ref{tab:xsum_nqswap}, we evaluated the performance of LLaMA-7B, LLaMA-3-8B-Instruct \cite{touvron2023bllama}, and Mistral-7B-v0.3 \cite{jiang2023mistral7b} models, and reported results in terms of fluency (ROUGE-L)\cite{lin2004rouge}, factual consistency (factKB)\cite{feng-etal-2023-factkb}), and semantic similarity (BERTScore-F1)\cite{zhang2019bertscore} on XSum, as well as Excact Match score on NQ-Swap. Across all models, HICD achieves a significant improvement in the factKB metric, while also maintaining consistent improvements in ROUGE-L and BERTScore-F1, indicating that the language quality and semantic coherence were not negatively impacted. The average scores across metrics further confirm that HICD achieves the most balanced and effective performance. HICD also shows competitive performance on NQ-Swap, achieving consistently favorable EM scores across all models.

\begin{table}[H]
\centering
\caption{Performance on Open-ended Generation Tasks}
\label{tab:xsum_nqswap}
\small
\renewcommand{\arraystretch}{1.15}
\begin{adjustbox}{width=\linewidth}
\begin{tabular}{lcccc|c}
\toprule
\multirow{2}{*}{\textbf{Model}} & \multicolumn{4}{c|}{\textbf{XSum}} & \textbf{NQ-Swap} \\
\cmidrule(lr){2-5} \cmidrule(lr){6-6}
& R-L ↑ & factKB ↑ & BERT-F1 ↑ & Avg ↑ & EM ↑ \\
\midrule
\textbf{LLaMA-7B}           & 17.80 & 47.21 & 63.76           & 42.9     & 56.25 \\
\quad + DoLA                & 17.84 & 47.15 & \textbf{64.13}  & 43.0     & 56.14 \\
\quad + CAD                 & 17.03 & 60.41 & 63.47  & 46.9 & 68.24 \\
\quad + HICD                & \textbf{17.98} & \textbf{61.25} & 62.17 & \textbf{47.1}     & \textbf{69.61} \\
\midrule
\textbf{LLaMA3-8B-Inst.}    & 19.71 & 46.53 & 65.34           & 43.9     & 58.74 \\
\quad + DoLA                & \textbf{19.84} & 47.68 & 65.11   & 44.2     & 58.86 \\
\quad + CAD                 & 18.73 & \textbf{63.21} & 64.98   & 49.0     & 72.51 \\
\quad + HICD                & 19.80 & 62.42 & \textbf{65.46}  & \textbf{49.2} & \textbf{72.64} \\
\midrule
\textbf{Mistral-7B-v0.3}    & 22.41 & 49.21 & 66.47           & 46.0     & 61.41 \\
\quad + DoLA                & \textbf{23.11} & 48.78 & 66.31   & 46.1     & 62.24 \\
\quad + CAD                 & 22.05 & 66.93 & 68.76           & 52.5     & 73.07 \\
\quad + HICD                & 22.68 & \textbf{67.21} & \textbf{68.84} & \textbf{52.9} & \textbf{73.73} \\
\bottomrule
\end{tabular}
\end{adjustbox}
\end{table}

\subsection{More Analysis}
\label{sec:more_analysis}
\textbf{Effect of inducing heads number on task performance of HICD.} We further analyze the relationship between the number of inducing heads and downstream task performance  with the LLaMA-7B model. The results, represented by all \textcolor{red}{red lines} in Figure~\ref{fig:induced_heads_analysis}, provide insight into this relationship.

For contextual faithfulness tasks, we adjust the number of \textit{Topk} heads to identify the optimal number of inducing heads. For the \textit{OpenBookQA} and \textit{RACE-High} tasks, a strong correlation between the number of inducing heads and accuracy. We attribute this to the strong dependence on the additional context provided in the datasets for making predictions. As a result, the inducing heads, which are crucial for capturing context relevant to the correctness of the model’s predictions, play an indispensable role. Increasing the number of inducing heads enables the model to generate more context-aware hallucinations, improving the effective of contrastive decoding and task performance. However, for \textit{HellaSwag} and \textit{RACE-Middle}, performance peaks at 30 inducing heads and decreases with further increases. We hypothesize that beyond a threshold, adding more inducing heads harms output, making contrastive decoding less effective and hindering performance. This is consistent with \cite{bansal-etal-2023-rethinking}, which observed that removing a significant percentage of attention heads greatly reduces model performance.

For factuality tasks, such as \textit{TruthfulQA}, a moderate correlation is observed between the number of inducing heads and various metrics, with Spearman correlations for \textit{MC1}, \textit{MC2} at 0.48, 0.65, respectively. However, the impact on performance is limited. For example, \textit{MC1} accuracy improves by just 1.8 points on \textit{TruthfulQA}, while accuracy for \textit{Wiki Factor} and \textit{News Factor} increases by 1.3\% and 3.5\%. We believe that hallucinations induced in factuality tasks are are less "contrast-effective" than in contextual tasks. As shown in Figure~\ref{fig:induced_heads_analysis}, hallucinations induced with fewer heads can even adversely affect contrastive decoding. Consequently, the hallucination mitigation effect of HICD is less prominent in factuality tasks, as the number of inducing heads changes. Nevertheless, in all experiments, HICD produces more accurate results than the baseline. Detailes analyses see Appendix \ref{sec:topk}.\vspace{1em}

\begin{table*}[t]
\centering
\caption
{Comparison of different hallucination-inducing methods across various evaluation tasks. Prompt-based, which uses a prompt to compel LLMs to provide fabricated information for contrast; PASTA-based, which employs attention steering to enhance the weights of low-importance tokens for inducing hallucinations; SH2-based, which prepends low-information tokens to redirect the model's attention toward unrelated context to induce hallucinations; Cut-based, which directly masks inducing heads to trigger hallucinations.}
\label{table2}
\resizebox{\textwidth}{!}{%
\begin{tabular}{lcccccccccccc}
\toprule
\multirow{2}{*}{\textbf{Methods}} & \textbf{Hellaswag} & \multicolumn{2}{c}{\textbf{Race}} & \multicolumn{2}{c}{\textbf{HaluEval-Sum}} & \textbf{OpenbookQA} & \multicolumn{3}{c}{\textbf{TruthfulQA}} & \multicolumn{2}{c}{\textbf{FACTOR}} \\
\cmidrule(lr){2-2} \cmidrule(lr){3-4} \cmidrule(lr){5-6} \cmidrule(lr){7-7} \cmidrule(lr){8-10} \cmidrule(lr){11-12}
& \textbf{Acc} & \textbf{Middle} & \textbf{High} & \textbf{Acc\_H} & \textbf{Acc\_A} & \textbf{Acc} & \textbf{MC1} & \textbf{MC2} & \textbf{MC3} & \textbf{WIKI} & \textbf{NEWS} \\
\midrule
Vanilla & 0.7760 & 0.5641 & 0.4339 & 18.94 & 26.06 & 0.5142 & 23.62 & 41.21 & 19.33 & 0.5855 & 0.5841 \\
+Prompt & 0.8025 & 0.5721 & 0.4454 & 21.61 & 25.82 & 0.5314 & 28.02 & 43.55 & 22.51 & 0.5841 & 0.5897 \\
+PASTA & 0.7859 & 0.5883 & 0.4408 & 26.57 & 29.25 & 0.5302 & 25.21 & 40.14 & 20.28 & 0.5955 & 0.5868 \\
+SH2 & 0.7971 & 0.5927 & 0.4436 & 25.96 & 26.01 & 0.5421 & \textbf{28.51} & 48.85 & 25.10 & \textbf{0.6279} & \textbf{0.6235} \\
+Cut & 0.8035 & 0.5829 & 0.4628 & 22.83 & \textbf{30.95} & 0.5402 & 25.09 & 51.83 & 26.33 & 0.6014 & 0.5932 \\
+HICD & \textbf{0.8423} & \textbf{0.5989} & \textbf{0.4668} & \textbf{27.15} & 27.21 & \textbf{0.5581} & 25.45 & \textbf{53.71} & \textbf{26.50} & 0.6058 & 0.6197 \\
\bottomrule
\end{tabular}%
}
\end{table*}

\begin{table*}[t]
\centering
\caption{In-domain and out-of-domain evaluation. Each row represents the performance of inducing heads, selected from different tasks, on a specific evaluation task. The best performance for each task is indicated in \textbf{bold}.}
\label{tab:metrics}
\small 
\resizebox{\textwidth}{!}{
\begin{tabular}{lccccccccc}
\toprule
\textbf{Metric} & \textbf{OpenbookQA} & \textbf{TruthfulQA} & \textbf{Race High} & \textbf{Halleswag} & \textbf{Factor News} & \textbf{Race Middle} & \textbf{Factor Wiki} & \textbf{HaluEval-Sum} & \textbf{Baseline}\\
\midrule
OpenbookQA & \textbf{0.558}   & 0.544   & 0.522   & 0.544   & 0.542   & 0.526   & 0.528  &0.542  & 0.514\\
TruthfulQA & 33.46   & \textbf{35.14}   & 32.30   & 34.90   & 34.11   & 33.96   & 31.20  & 33.85 &28.05\\
Race High  & 0.453   & 0.457   & \textbf{0.469}   & 0.454   & 0.451  & 0.449   & 0.445  & 0.458 & 0.434\\
Halleswag  & 0.813   & 0.827   & 0.804   & \textbf{0.842}   & 0.808  & 0.834   & 0.809  & 0.808 & 0.776\\
Factor News     & 0.585   & 0.588   & 0.575 & 0.583 & \textbf{0.619} & 0.589 & 0.571 & 0.581 & 0.584\\
Race Middle & 0.583  & 0.588   & 0.568   & 0.596   & 0.563  & \textbf{0.598}   & 0.572  &0.581 &0.564\\
Factor Wiki     & 0.588   & 0.583   & 0.576 & 0.581 & 0.572 & 0.584 & \textbf{0.605} & 0.590 & 0.585\\
HaluEval-Sum    & 24.85  & 26.07  & 24.83  & 20.61   & 23.36  & 27.31   &24.05   & \textbf{35.22} & 22.51\\
\bottomrule
\end{tabular}%
}\vspace{-0.3cm}
\end{table*}

\noindent\textbf{Comparison with Other Hallucination-Inducing Methods}. HICD demonstrates an ability to induce more "contrast-effective" hallucinations compared to other methods. We compare HICD with the following methods, as detailed in Appendix \ref{sec: E appendix}:
\begin{itemize}
    \setlength{\itemsep}{0pt}
    \setlength{\parskip}{0pt}
    \setlength{\topsep}{0pt}
    \setlength{\leftmargini}{0pt}
    \item \hypertarget{sec:prompt}{\textbf{Prompt-based}}: A prompt is used to force LLMs to generate fabricated information to induce hallucinations.
    \item \hypertarget{sec:sh2}{\textbf{SH2-based}}: Low-information tokens are prepended to the context to shift the model's attention to unrelated content to induce hallucinations\cite{kai-etal-2024-sh2}.
    \item \hypertarget{sec:pasta}{\textbf{PASTA-based}}: Attention steering is applied by increasing the attention weights of low-importance tokens to induce hallucination\cite{zhang2023tell}.
    \item \hypertarget{sec:cut}{\textbf{Cut-based}}: Inducing heads are directly masked to trigger hallucinations.
\end{itemize}

As shown in Figure~\ref{fig:induced_heads_analysis}, in most tasks, the Cut-based method (\textcolor{blue}{blue lines}) exhibits a weaker ability to mitigate hallucinations at the optimal number of inducing head compared to the Ave-based approach HICD (\textcolor{red}{red lines}). From the results in Table~\ref{table2}, HICD consistently outperforms both the Prompt-based and PASTA-based method in most datasets. This is especially evident in contextual faithfulness tasks, where HICD achieves the best overall performance.

\begin{figure}[h]
\centering
\includegraphics[width=\columnwidth]{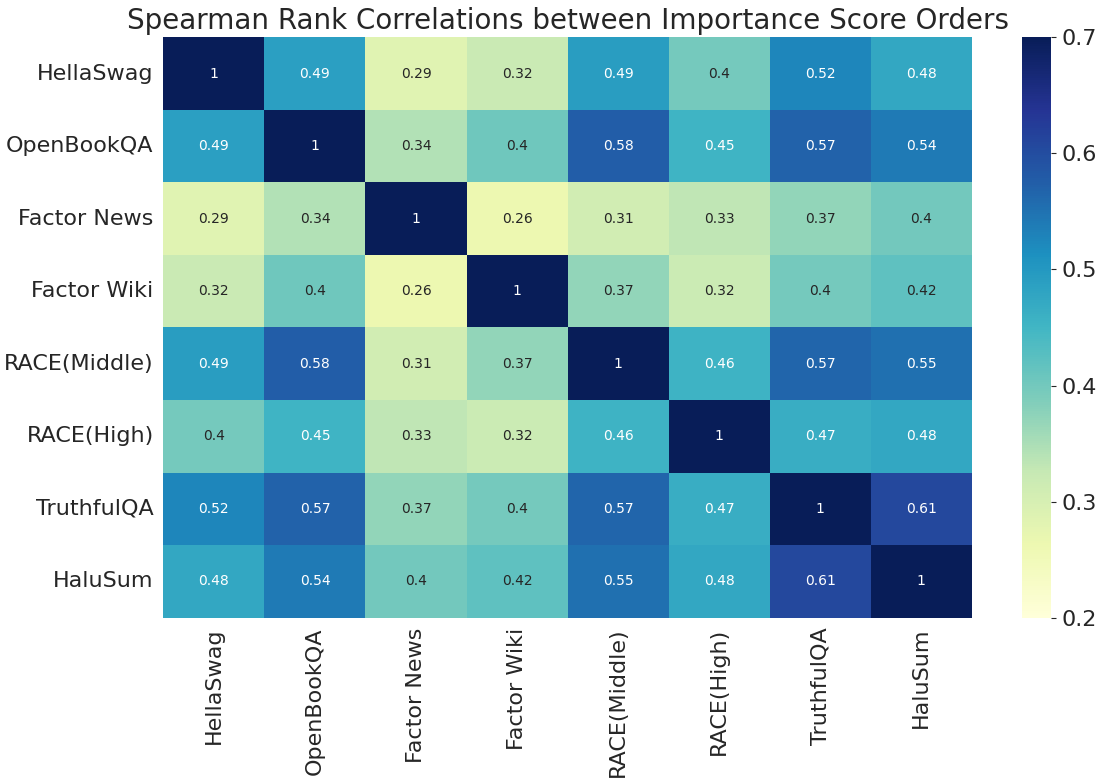} 
\caption{Spearman correlation coefficients for inducing heads score ranking across different tasks. Higher correlation coefficients indicate that the inducing heads selected more similarly. }
\label{fig:spearman}
\end{figure}
\setlength{\textfloatsep}{7pt}

Although SH2-based method for inducing hallucination outperforms HICD on specific factuality tasks, such as the \textit{TruthfulQA} in MC1 metric and the \textit{FACTOR} datasets, the overall results indicate that HICD has a greater potential for inducing "contrast-effective" hallucinations. This advantage makes HICD particularly effective in mitigating hallucinations while maintaining superior performance in a wide range of evaluation tasks.\vspace{1em}

\noindent\textbf{In-domain and Out-of-domain Inducing Head Evaluation.} We evaluate the performance of in-domain and out-of-domain inducing head selection method, with the results presented in Table~\ref{tab:metrics}. For the in-domain setup, the inducing heads are selected using the specific task dataset and evaluated on the same task. For the out-of-domain setup, the inducing heads are selected from a task dataset and tested on different tasks.

The highest performance is consistently obtained from in-domain inducing heads. This demonstrates that task-relevant, in-domain head selection outperforms out-of-domain selection methods across all datasets, significantly improving model performance. Moreover, the results for out-of-domain inducing heads are generally better than baseline methods, indicating that the HICD approach exhibits a certain degree of generalizability across different datasets and tasks. \vspace{1em}

\noindent\textbf{The performance of out-of-domain inducing heads is related to the correlation between in-domain and out-of-domain heads rankings.} As the correlation between out-of-domain and in-domain inducing heads increases, their performance becomes more similar, with results presented in Figure~\ref{fig:spearman}. For example, the inducing heads from Race Middle, TruthfulQA, and HaluEval-Sum exhibit relatively high ranking similarity with OpenBookQA. Therefore, the out-of-domain heads from these tasks show performance that is notably closer to the in-domain OpenBookQA heads compared to other out-of-domain heads, as seen in Table~\ref{tab:metrics}. Similarly, the Factor (News and Wiki) tasks exhibit relatively lower Spearman correlation with other tasks, leading to similar performance among the Factor's out-of-domain heads, which shows a significant gap in performance compared to in-domain heads. See details in Appendix \ref{sec: F appendix}.\vspace{1em}

\begin{figure}[t]
\centering
\includegraphics[width=\columnwidth]{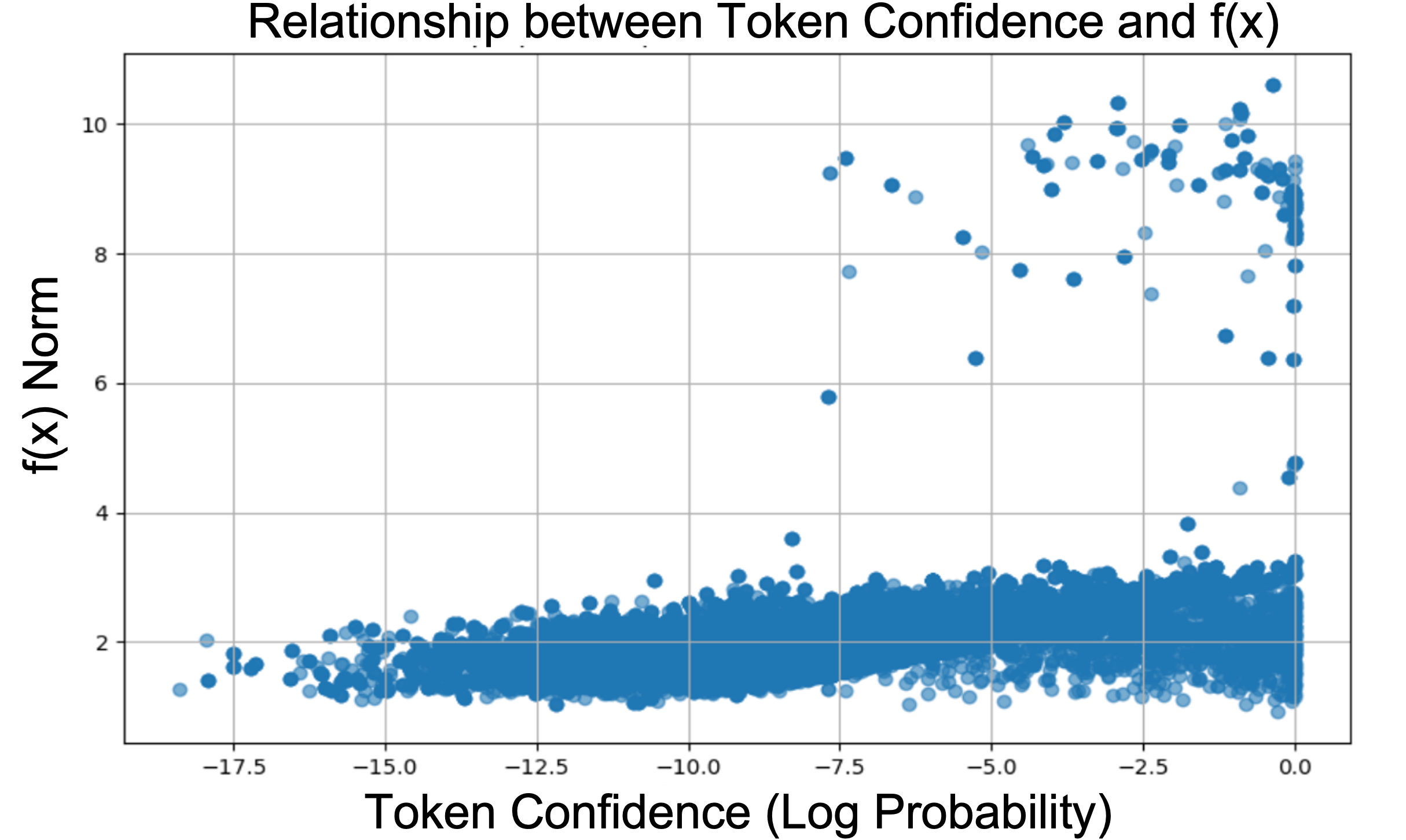}  
\caption{Visualization of the relationship between token confidence and the norm \( f(x) \), where a subset of high-confidence tokens corresponds to higher \( f(x) \).}
\label{fig:confidence}
\end{figure}

\noindent\textbf{Analysis of Attention Map Averaging vs. Head Cutting in Inducing Effective Hallucinations}. The attention mechanism transforms each input vector \( x \) into a norm \( f(x) \), calculates the attention weights \( \alpha \), then computes the output \( \alpha f(x) \). Compared to \( \alpha \), the \( f(x) \) plays the dominant role in controlling the attention of the high-frequency and low-information tokens \cite{kobayashi-etal-2020-attention}. Besides, a higher token confidence corresponds to a lower information content \cite{kai-etal-2024-sh2}.

\begin{figure}[t]
\centering
\includegraphics[width=\columnwidth]{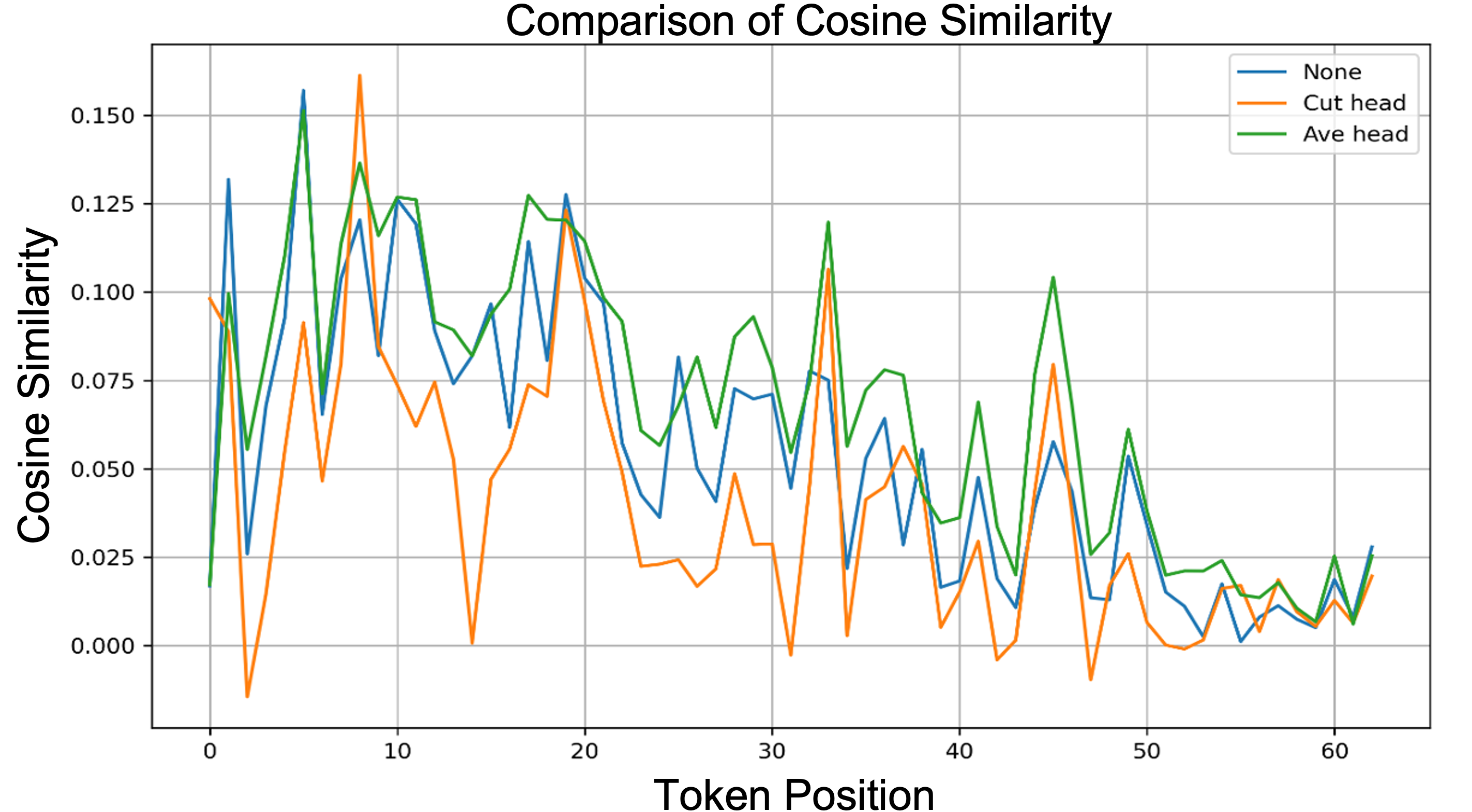} 
\caption{Cosine similarity of the output norms \( ||f(x)|| \) and \( ||\alpha f(x)|| \)(attention output) at different token positions under the methods: None, Cut Head, and Ave Head. Ave Head shows a higher similarity, allowing \( ||f(x)|| \) to dominate the final attention values.}
\label{fig:similarity}
\vspace{-0.2cm}
\end{figure}

Building on these intuitions, we analyze the relationship between token confidence and the norm \( f(x) \), as illustrated in Figure~\ref{fig:confidence}. Most tokens exhibit low \( f(x) \) values, but a subset of high-confidence, low-information tokens corresponds to higher \( f(x) \) values. We hypothesize this strengthens the final attention values at the positions of low-information tokens. Figure~\ref{fig:similarity} compares the cosine similarity of the \( ||f(x)|| \) and  \( ||\alpha f(x)|| \)(attention output) at different token positions across three methods. As shown, Ave Head results in higher similarity between \( ||f(x)|| \) and \( ||\alpha f(x)|| \) than  the others, increasing the dominance of \( ||f(x)|| \) in determining the final attention values. Thus, HICD applies attention map averaging makes \( \alpha \) uniform across all positions, with the final attention determined by \( f(x) \). Higher values of \( f(x) \), dominated by low-information tokens, exert a greater influence.

\begin{figure}[t]
\centering
\includegraphics[width=\columnwidth]{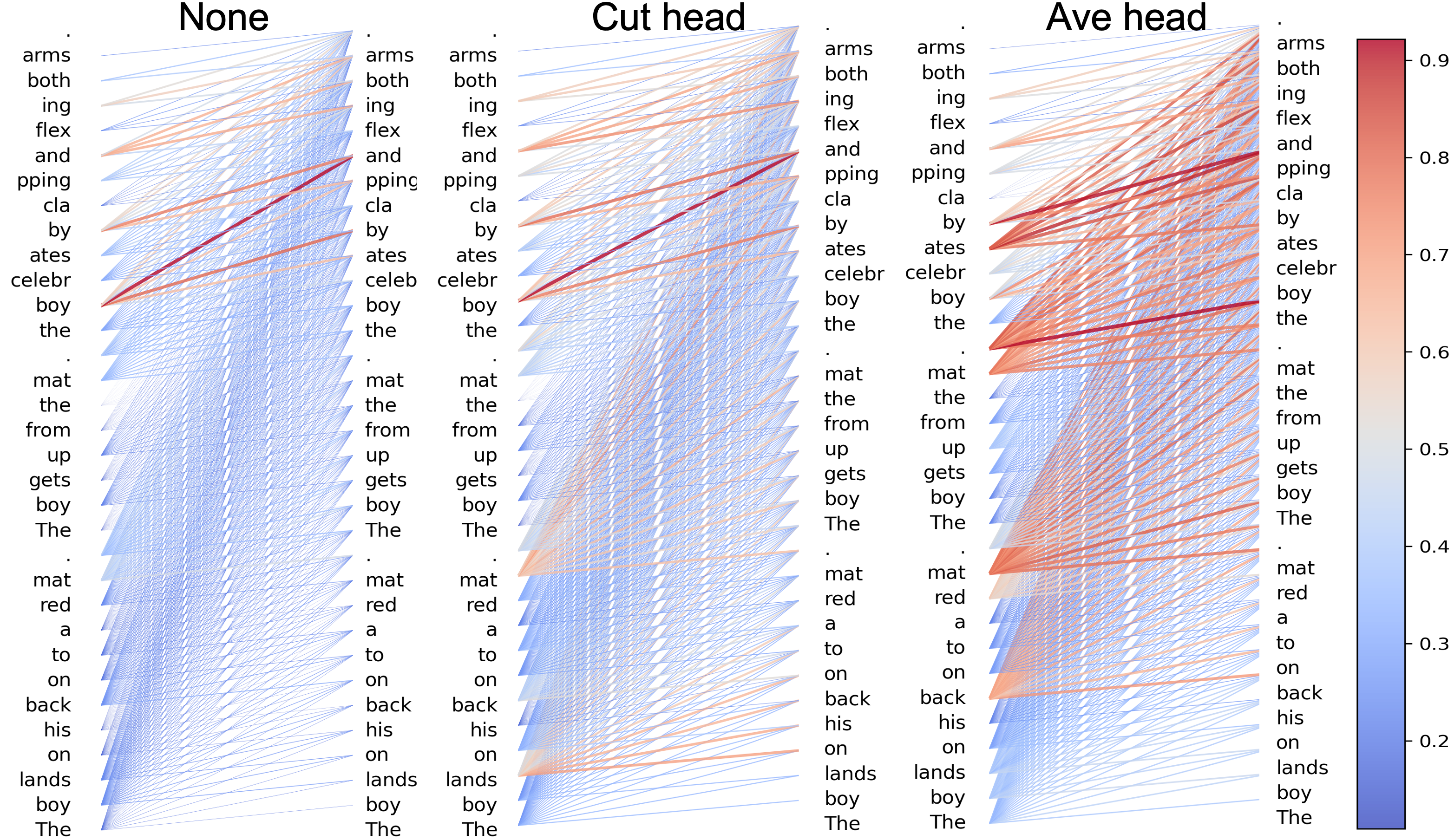} 
\caption{Visualization of the information flow, Ave head increases the importance of information flow from more tokens, leading to spread-out attention distribution and more plausible hallucinations.}
\label{fig:saliency}
\end{figure}

To further illustrate the impact of attention averaging on model's outputs. Based on \cite{wang-etal-2023-label}, we visualize the information flow in Figure~\ref{fig:saliency}. Compared to other methods, Ave Head increases the importance of information flow from more tokens to the token being predicted, making the model consider the impact of other irrelevant low-information tokens. This makes the hallucinated outputs seem more plausible and meaningful.

\section{Conclusion}

In this paper, HICD are introduced to induce hallucinations on inducing heads for contrastive decoding to mitigate hallucinations. Experiments on several tasks show that HICD outperforms existing methods in contextual tasks and achieves competitive results in factual consistency tasks. We also find that selecting task-relevant inducing heads improves performance compared to out-of-domain selections. And attention averaging induces more contrast-effective hallucinations compared to other methods. Our work opens new directions for hallucination induction and mitigation, providing a promising strategy to reduce hallucinations and enhance LLM robustness across tasks.

\section{Limitations}
The HICD method shows strong improvements in hallucination mitigation, but it has several limitations. First, its effectiveness depends on task-relevant induced head selection, which may not generalize well to all tasks, especially those underrepresented in training data.
Second, attention map averaging for hallucination induction can be computationally expensive, particularly for larger models and datasets, making scalability a concern for real-time or resource-limited applications. Lastly, the method’s performance relies on the quality of adversarial data, and future work should explore how different adversarial data construction methods impact performance across various tasks and domains.

\section{Acknowledgements}
This work was supported by the National Natural Science Foundation of China (NSFC) under grant no. 12475196 and 12373113.

\bibliography{custom}

\clearpage

\appendix

\section{Experimental Setup Details}
\label{sec: A appendix}

\subsection{Datasets and Metrics}

\textbf{1) Faithfulness Evaluation}

For faithfulness evaluation, we use the following tasks:

\begin{itemize}
\item \textbf{Context Completion (HellaSwag):} HellaSwag \cite{zellers2019hellaswag} is a dataset designed to evaluate the ability of a model to predict the next sentence based on context. It contains multiple-choice questions that require the model to select the most plausible continuation of a given context. The task tests how well the model maintains context coherence and handles commonsense reasoning. We use the validation split of HellaSwag, which contains 10,042 examples. The dataset can be accessed at: \url{https://huggingface.co/datasets/Rowan/hellaswag}.

\item \textbf{Reading Comprehension (RACE):} RACE \cite{lai-etal-2017-race} is a reading comprehension dataset that contains two subsets: RACE-H (high school) and RACE-M (middle school). The dataset consists of questions based on passages, requiring the model to select the correct answer. RACE tests the model’s ability to understand and reason about the context of longer text. We use the test split of RACE, with RACE-H containing 3,498 examples and RACE-M containing 1,436 examples. The dataset is available at: \url{https://huggingface.co/datasets/ehovy/race}.

\item \textbf{Question Answering (OpenBookQA):} OpenBookQA \cite{mihaylov-etal-2018-suit} is a dataset designed to evaluate a model's ability to answer scientific questions. It consists of two subsets: main and additional. The additional subset provides a 'fact1' field as a reference context, which contains core scientific facts related to the question. In our evaluation, we use the additional subset and treat 'fact1' as the contextual input for the model. We use the test split of the additional subset, which contains 500 examples. This task assesses the model’s ability to recall and apply scientific knowledge in a reasoning context. The dataset is available at: \url{https://huggingface.co/datasets/allenai/openbookqa}.

\end{itemize}

\textbf{2) Knowledge Hallucination Evaluation}

To assess the extent of hallucinations generated by the model, we utilize the following task:

\begin{itemize}
  \item \textbf{HaluEval-Sum:} HaluEval \cite{li-etal-2023-halueval} is used to evaluate hallucinations in summaries generated by the model. This dataset includes 10,000 samples, where each sample consists of a document, a hallucinated summary, and a correct summary. The task involves determining whether a summary contains factual inconsistencies or hallucinations. The performance of the model is evaluated using two metrics:
  The dataset can be accessed at: 
    \begin{itemize}
    \item Arithmetic-mean accuracy (Acc-A): The mean accuracy for both hallucinated and correct summaries.
    \item Harmonic-mean accuracy (Acc-H): The harmonic mean of the accuracy for hallucinated and correct summaries. Acc-H provides a more balanced view, penalizing imbalances between the two types of summaries.
  \end{itemize}
  The dataset can be accessed at: \url{https://github.com/RUCAIBox/HaluEval/blob/main/data/summarization_data.json}.
\end{itemize}

\textbf{3) Factuality Evaluation}

For evaluating factual consistency, we use the following datasets:

\begin{itemize}
  \item \textbf{TruthfulQA:} TruthfulQA \cite{lin-etal-2022-truthfulqa} is a dataset designed to test the truthfulness of language models. It consists of multiple-choice questions where the model must select the correct answer from a set of options. The dataset includes three metrics for evaluating the model's truthfulness. We use the validation split of the multiple-choice subset, which contains 817 examples. The dataset is available at:\url{https://huggingface.co/datasets/truthfulqa/truthful_qa/viewer/multiple_choice}.

  \item \textbf{FACTOR (Wiki and News):} The FACTOR dataset \cite{muhlgay-etal-2024-generating} focuses on factual consistency, requiring the model to select the correct completion of a text from factual and non-factual alternatives. It includes two subsets: Wiki-FACTOR and News-FACTOR, with 2,994 and 1,036 examples, respectively. The task tests the model’s ability to generate factually accurate outputs. The dataset is available at:\url{https://github.com/AI21Labs/factor/tree/main/data}.
\end{itemize}

\subsection{Models and Baselines}

We conduct our experiments with the Llama family of models \cite{touvron2023allama,touvron2023bllama}. The following baseline methods are used for comparison:

\begin{itemize}
  \item \textbf{Greedy Decoding:} This baseline method selects the next token greedily by choosing the one with the highest probability at each step. It is the simplest form of decoding and serves as a baseline for comparison with more advanced methods.
  \item \textbf{DoLa:} DoLa \cite{chuang2024dola} is a contrastive decoding method that attempts to reduce hallucinations by contrasting the output distributions of different layers of the model. This method aims to enhance the factuality of the generated text by comparing the outputs from various layers. The method code is available at: \url{https://github.com/voidism/DoLa}.
  \item \textbf{Context-Aware Decoding (CAD):} CAD \cite{shi-etal-2024-trusting} is a variant of contrastive decoding that involves two models: the first model, which has access to the full context during decoding, and the second model, which is the same architecture but lacks access to the additional context. By contrasting their outputs, CAD amplifies the difference in performance when the model has context, helping it focus more on the provided context. This improves the model’s faithfulness, particularly when the context introduces new or contradictory information. The method code is available at: \url{https://github.com/xhan77/context-aware-decoding}.
  \item \textbf{Contrastive Decoding (CD):} CD \cite{li-etal-2023-contrastive} is a well-established contrastive decoding method that contrasts the token distributions of models with different parameter scales. This approach aims to reduce hallucinations by comparing the outputs of smaller(7b) models with larger(13b), more powerful models. The method code is available at: \url{https://github.com/XiangLi1999/ContrastiveDecoding}.
\end{itemize}

\subsection{Computational Resources and Software Libraries}

\textbf{Model and Computational Resources.}  
All experiments were conducted using the Llama\footnote{\url{https://huggingface.co/huggyllama/llama-7b}} , Alpaca\footnote{\url{https://huggingface.co/wxjiao/alpaca-7b}} and Llama2\footnote{\url{https://huggingface.co/meta-llama/Llama-2-7b-hf}}  models, both of which have 7 billion parameters. We test the inference experiments executed on a single Tesla V100 GPU (32GB) without GPU parallelism. The approximate runtime for inference on different datasets show in Table~\ref{tab:inference_time}.

\begin{table}[t]
\centering
\small
\caption{Inference time for different datasets using a single Tesla V100 (32GB) GPU.}
\label{tab:inference_time}
\resizebox{\linewidth}{!}{%
\begin{tabular}{lcc}
\toprule
\textbf{Dataset} & \textbf{Number of Examples} & \textbf{Inference Time} \\
\midrule
HellaSwag & 10,042 & 82 m \\
RACE-M & 1,436 & 14 m \\
RACE-H & 3,498 & 52 m \\
OpenBookQA & 500 & 3 m \\
TruthfulQA & 817 & 18 m \\
FACTOR-Wiki & 2,994 & 40 m \\
FACTOR-News & 1,036 & 13 m \\
HaluEval-Sum & 10,000 & 15 h \\
\bottomrule
\end{tabular}%
}
\end{table}

\noindent\textbf{Software and Implementations.}  
We utilized PyTorch\footnote{\url{https://github.com/pytorch/pytorch}} , and Transformers\footnote{\url{https://github.com/huggingface/transformers}} . The Transformers library was modified to support head masking and attention map averaging. Additionally, Baukit\footnote{\url{https://github.com/davidbau/baukit}} and lm-evaluation-harness\footnote{\url{https://github.com/EleutherAI/lm-evaluation-harness}} were used in our implementation.The modifications to the Transformers library primarily focused on adding a head\_mask attribute to control inducing heads and implementing attention map averaging .

This setup ensures that our experimental results are reproducible and that sufficient computational resources were allocated for evaluating model performance across multiple benchmarks.

For the reported experimental results, we set the random seed to 42 for all runs to ensure reproducibility. The results presented are based on the maximum performance observed across multiple runs with the same seed. Specifically, for each dataset, we ran experiments using a fixed seed and report the highest accuracy obtained across different validation or test splits. We emphasize that these results represent the best-performing configurations under this particular seed setting. Additionally, while the results are based on a single random seed for consistency, future work could benefit from running experiments across multiple seeds to better assess the stability and reliability of the model's performance.

\subsection{Identification of Inducing Heads}\label{sec:Experimental Setup Details}

In this subsection, we focus on how to construct adversarial data using incorrect answer options from the original dataset. By utilizing these adversarial samples, we calculate the importance scores for attention heads that are crucial for predicting incorrect answers, referred to as "wrong heads." This process allows us to evaluate the impact of these heads on the model's performance in generating erroneous outputs. We directly utilize the other answer choices in the dataset (which are not the correct answer) and treat them as adversarial labels. Using the gradient-based importance scoring method, we compute the importance scores for each attention head that influences the model’s decision towards a wrong answer. The higher the score, the more important that head is in contributing to the model’s incorrect response. We then compute the average importance score for the heads corresponding to all adversarially constructed data and use this average score as the final importance score for the "wrong heads."

In parallel, we also compute the importance scores for "right heads" using the original correct answers. These heads are critical in generating correct outputs, and their scores provide insights into the attention heads responsible for guiding the model toward accurate decisions.

The final inducing heads score is determined by combining the scores of both "right" and "wrong" heads. This allows us to identify which heads are most influential in guiding the model’s decisions towards  outputs. The optimal number of inducing heads is chosen based on the combined importance scores, as detailed in Section \ref{sec:experiments}. 

\section{Parameter Settings Analysis and Hyperparameter Tuning}
\label{sec: B appendix}

Table \ref{table2} has demonstrated the impact of selecting top-k inducing heads on model performance across different tasks. In this section, we provide a detailed account of the parameter configurations used in our experiments, including the hyperparameter values and their corresponding evaluation results. As shown in Table \ref{tab:ablation_alpha_effect}, we investigate the effect of the hyperparameter \(\alpha\) on model performance while 

\begin{table}[H]  
\centering
\caption{Ablation study showing the effect of Alpha on the evaluation results, with fixed Scale and Top-k.}
\label{tab:ablation_alpha_effect}
\small
\resizebox{\columnwidth}{!}{
\begin{tabularx}{\columnwidth}{lXXX}
\toprule
\textbf{Task} & \textbf{Alpha} & \textbf{Scale, Top-k} & \textbf{Evaluation} \\
\midrule
\textbf{HellaSwag} & 0.7  & 20, 30  & 0.8325 \\
                   & 0.9  & 20, 30  & 0.8379 \\
                   & 1.1  & 20, 30  & 0.8422 \\
                   & 1.3  & 20, 30  & 0.8421 \\
                   & 1.5  & 20, 30  & 0.8413 \\
                   & 1.7  & 20, 30  & 0.8424 \\
\midrule
\textbf{Race Middle} & 0.5  & 10, 30  & 0.5974 \\
                     & 0.7  & 10, 30  & 0.5988 \\
                     & 0.9  & 10, 30  & 0.5968 \\
                     & 1.1  & 10, 30  & 0.5912 \\
                     & 1.3  & 10, 30  & 0.5863 \\
                     & 1.5  & 10, 30  & 0.5856 \\
\midrule
\textbf{Race High} & 0.5  & 50, 70  & 0.4594 \\
                   & 0.7  & 50, 70  & 0.4608 \\
                   & 0.9  & 50, 70  & 0.4628 \\
                   & 1.1  & 50, 70  & 0.4599 \\
                   & 1.3  & 50, 70  & 0.4637 \\
\midrule
\textbf{OpenbookQA} & 0.6  & 1, 70  & 0.544 \\
                    & 0.8  & 1, 70  & 0.558 \\
                    & 1.0  & 1, 70  & 0.542 \\
                    & 1.2  & 1, 70  & 0.538 \\
                    & 1.4  & 1, 70  & 0.546 \\
\midrule
\textbf{TruthfulQA} & -1.0  & 10, 70  & 0.2386 0.4573 0.2387\\
                    & -3.0  & 10, 70  & 0.2533 0.4852 0.2589\\
                    & -5.0  & 10, 70  & 0.2533 0.5105 0.2641\\
                    & -6.0  & 10, 70  & 0.2545 0.5339 0.2644\\
                    & -7.0  & 10, 70  & 0.2521 0.5187 0.2638\\
\midrule
\textbf{Factor News} & 0.3  & 10, 70  & 0.5984 \\
                     & 0.38 & 10, 70  & 0.6197 \\
                     & 0.42 & 10, 70  & 0.6003 \\
                     & 0.44 & 10, 70  & 0.6004 \\
                     & 0.5  & 10, 70  & 0.5984 \\
\midrule
\textbf{Factor Wiki} & 0.38 & 20, 70  & 0.5935 \\
                     & 0.5  & 20, 70  & 0.6058 \\
                     & 0.8  & 20, 70  & 0.5931 \\
                     & 1.0  & 20, 70  & 0.5945 \\
                     & 1.3  & 20, 70  & 0.5902 \\
\midrule
\textbf{HaluEval-Sum} & 0.3  & 20, 30  & 25.31 26.50\\
                      & 0.5  & 20, 30  & 26.01 26.70\\
                      & 0.7  & 20, 30  & 26.44 26.75\\
                      & 0.9  & 20, 30  & 27.15 27.25\\
                      & 1.1  & 20, 30  & 27.02 27.15\\
\bottomrule
\end{tabularx}
}
\end{table}
\noindent keeping Scale \( s\) and Top-k fixed. Similarly, in Table \ref{tab:ablation_scales_effect}, we explore how Scale \(\ s\) influences performance while keeping \(\alpha\) and Top-k fixed.

\subsection{Effect of \(\alpha\) (Alpha)}
The \(\alpha\) parameter controls the relative weighting between the original model and the hallucination-induced model during contrastive decoding (Equation \ref{eq:contrastive_decoding}). A higher \(\alpha\) amplifies the suppression of hallucinated outputs, while a lower \(\alpha\) allows more hallucination-driven tokens. 

As seen in Table \ref{tab:ablation_alpha_effect}, the effect of \(\alpha\) on model performance varies by task. For example, in HellaSwag, increasing \(\alpha\) from 0.7 to 1.1 leads to a steady improvement, but further increases provide diminishing returns, with performance stabilizing around \(\alpha = 1.3\). A similar trend is observed in Race Middle, where performance peaks at \(\alpha = 0.7\), after which further increases cause a decline. In contrast, for TruthfulQA, a significantly larger negative value of \(\alpha = -6.0\) provides optimal performance. In our experiments, we found that a negative \(\alpha\) forces the model to prioritize the hallucinated outputs generated by the induced model over the original model's outputs. For TruthfulQA, this leads to a more effective combination of the original model and the inducing model outputs, improving the overall performance. For Factor tasks, the effect of changing \(\alpha\) was less pronounced, which may be due to the inducing hallucinations not being as contrast-effective in these tasks compared to others. This suggests that hallucinations induced in Factor tasks do not contribute as effectively to contrastive decoding, leading to relatively smaller performance improvements when adjusting \(\alpha\).

\subsection{Effect of Scale Parameter}
The Scale parameter \(s\) determines the weight of the discrepancy correction factor applied during inducing head selection (Equation \ref{eq:gradient_score}). It adjusts how much difference in importance scores between correct and incorrect outputs influences the final inducing head score.

As shown in Table \ref{tab:ablation_scales_effect}, the optimal Scale value varies in different tasks, with each task exhibiting a distinct best value for Scale \(s\). Scale \(s\) effectively adjusts the importance scores of the inducing heads, which in turn influences the selection of more contrast-effective inducing heads. For instance, in the HellaSwag task, the performance peaks at \(\ s = 20\), while in Race Middle, the best performance is achieved at \(\ s = 10\). Compared to \(\alpha\), the effect of \(s\) on performance is relatively subtle, as it primarily changes the scores used to 

\begin{table}[H]
\centering
\caption{Ablation study showing the effect of Scale on the evaluation results, with fixed Alpha and Top-k.}
\label{tab:ablation_scales_effect}
\small
\resizebox{\columnwidth}{!}{
\begin{tabularx}{\columnwidth}{lXXX}
\toprule
\textbf{Task} & \textbf{Scale} & \textbf{Alpha, Top-k} & \textbf{Evaluation} \\
\midrule
\textbf{HellaSwag} & 10  & 1.1, 30  & 0.8326 \\
                   & 20  & 1.1, 30  & 0.8422 \\
                   & 30  & 1.1, 30  & 0.7491 \\
                   & 50  & 1.1, 30  & 0.7422 \\
                   & 70  & 1.1, 30  & 0.7310 \\
                   & 100 & 1.1, 30  & 0.7367 \\
\midrule
\textbf{Race Middle} & 1  & 0.7, 30  & 0.5968 \\
                     & 10 & 0.7, 30  & 0.5988 \\
                     & 20 & 0.7, 30  & 0.5842 \\
                     & 50 & 0.7, 30  & 0.5842 \\
                     & 70 & 0.7, 30  & 0.5815 \\
                     & 100 & 0.7, 30  & 0.5864 \\
\midrule
\textbf{Race High} & 1  & 1.3, 70  & 0.4603 \\
                   & 10  & 1.3, 70  & 0.4643 \\
                   & 20  & 1.3, 70  & 0.4631 \\
                   & 50  & 1.3, 70  & 0.4651 \\
                   & 70  & 1.3, 70  & 0.4634 \\
                   & 100  & 1.3, 70  & 0.4668 \\
\midrule
\textbf{OpenbookQA} & 1  & 0.8, 70  & 0.5441 \\
                    & 10 & 0.8, 70  & 0.5582 \\
                    & 20 & 0.8, 70  & 0.5380 \\
                    & 30 & 0.8, 70  & 0.5364 \\
                    & 50 & 0.8, 70  & 0.5307 \\
\midrule
\textbf{TruthfulQA} & 1 & -6, 70  & 0.2264  0.5019  0.2501 \\
                    & 10 & -6, 70  & 0.2545  0.5339  0.2644 \\
                    & 30 & -6, 70  & 0.2337  0.5177  0.2538 \\
                    & 50 & -6, 70  & 0.2423  0.5209  0.2613 \\
                    & 100 & -6, 70  & 0.2386  0.5188  0.2588 \\
\midrule
\textbf{Factor News} & 1 & 0.38, 70  & 0.5917 \\
                     & 10 & 0.38, 70  & 0.6197 \\
                     & 30 & 0.38, 70  & 0.5782 \\
                     & 50 & 0.38, 70  & 0.5782 \\
                     & 70 & 0.38, 70  & 0.5839 \\
\midrule
\textbf{Factor Wiki} & 1 & 0.5, 70  & 0.5961 \\
                     & 10 & 0.5, 70  & 0.5962 \\
                     & 30 & 0.5, 70  & 0.6058 \\
                     & 50 & 0.5, 70  & 0.5961 \\
                     & 70 & 0.5, 70  & 0.5958 \\
\midrule
\textbf{HaluEval-Sum} & 20 & 0.9, 30  & 27.15  27.25 \\
                      & 100 & 0.9, 30  & 25.45  25.70 \\
\bottomrule
\end{tabularx}
}
\end{table}

 \noindent select inducing heads rather than directly impacting the final output. This indicates that while \(s\) changes the hallucination induction process by altering the selection of inducing heads, it does not drastically impact the model's overall contrastive decoding performance.

\subsection{Effect of Inducing Head Selection (Top-k)}
\label{sec:topk}
The number of inducing heads (\textit{Top-k}) plays a crucial role in determining the extent of hallucination induction and contrastive decoding effectiveness. As observed in Figure \ref{fig:induced_heads_analysis}, different tasks achieve peak performance at different Top-k values. In HellaSwag selecting 30 inducing heads yields optimal results, whereas OpenBookQA performs best with 70 inducing heads. This suggests that different tasks have different sensitivities to hallucination induction, and optimal Top-k values should be determined based on task-relevant characteristics rather than a fixed number across all tasks.

As shown in Table \ref{tab:topk_effect}, selecting an appropriate Top-k value improves the performance of the model on various tasks. For example, in \textit{HellaSwag}, selecting 30 inducing heads yields the highest accuracy of 0.8423, while in \textit{RACE High}, the optimal number of inducing heads is 70, resulting in an accuracy of 0.4637. In \textit{OpenBookQA}, selecting 70 inducing heads also provides the best performance with an accuracy of 0.558. From the extent of the impact of varying the selected Top-k on performance,we confirm that Top-k selection plays a crucial role in optimizing the model's performance by effectively inducing hallucinations for contrastive decoding.

In tasks like \textit{TruthfulQA}, the optimal Top-k selection varies depending on the evaluation metric. For instance, the MC1, MC2, and MC3 scores achieve peak values at Top-k = 70, which suggests that the inducing heads selected at this value help the model focus on the right hallucinations to improve factual correctness across the multiple-choice questions. Similarly, for \textit{Race Middle}, the performance improves as Top-k increases, with 30 inducing heads yielding the best results. However, increasing Top-k further leads to diminishing returns, emphasizing the importance of selecting an optimal number of heads for each task.

These findings suggest that while increasing the number of inducing heads can enhance performance up to a certain point, there exists an optimal threshold beyond which adding more heads does not yield further benefits. In fact, as the number of inducing heads continues to increase, the hallucinations inducing become less contrast-effective and can even lead to worse performance compared to the original model outputs. This indicates that hallucination induction should be balanced. An excessive number of inducing heads can introduce noise, diluting the effectiveness of the contrastive decoding process. Therefore, it is crucial to fine-tune Top-k based on task-relevant characteristics to maintain the effectiveness of hallucination induction without surpassing the point of diminishing returns. The results with Llama2-7b are shown in Table ~\ref{llama2_topk}.

\subsection{Final Hyperparameter Selection}
After extensive tuning, we summarize the optimal hyperparameter configurations in Table \ref{tab:final_hyperparams}. These values were selected based on maximizing performance across all evaluation metrics while ensuring stable and reliable contrastive decoding.

Overall, our analysis highlights the importance of careful hyperparameter tuning in balancing hallucination induction and mitigation. The results demonstrate that an appropriate combination of \(\alpha\), Scale, and Top-k effectively enhances model robustness in contrastive decoding, with different tasks requiring distinct configurations to achieve optimal performance.
\begin{table}[t]
    \centering
    \caption{Final hyperparameter configurations for each task, optimized based on performance across evaluation metrics.}
    \label{tab:final_hyperparams}
    \small
    \begin{tabularx}{\columnwidth}{lXXX}
        \toprule
        \textbf{Task} & \textbf{Alpha} & \textbf{Scale} & \textbf{Top-k} \\
        \midrule
        HellaSwag     & 1.1  & 20  & 30  \\
        Race Middle  & 0.7  & 10  & 30  \\
        Race High    & 1.3  & 50  & 70  \\
        OpenBookQA   & 0.8  & 1  & 70  \\
        TruthfulQA   & -6.0  & 10  & 70  \\
        Factor News  & 0.38 & 10  & 70  \\
        Factor Wiki  & 0.5  & 20  & 70  \\
        HaluEval-Sum & 0.9  & 20  & 30  \\
        \bottomrule
    \end{tabularx}
\end{table}

\section{Additional Results and Analysis}
\label{sec: C appendix}

\subsection{Head Pruning Method in Our Experiments}
\label{sec:head_pruning}

In our experiments, the head-pruning method is implemented by directly setting the inducing heads to be inactive. This process effectively "prunes" the selected heads by disabling them in the attention mechanism. Specifically, this involves setting the attention values of the chosen inducing heads to zero, which ensures that these heads do not contribute to the final output. As a result, the output from the pruned heads is excluded from the overall attention computation, effectively simulating a head pruning. This method serves as a baseline for comparison with the HICD method, where hallucinations are induced by averaging the attention maps of selected heads.

\begin{table}[H]
\centering
\caption{Ablation study showing the effect of Top-k inducing heads on model performance across various tasks.}
\label{tab:topk_effect}
\small
\resizebox{\columnwidth}{!}{
\begin{tabularx}{\columnwidth}{lcccccc}
\toprule
\textbf{Task} & \textbf{Top-k} & \textbf{Acc / MC}  \\
\midrule
\textbf{Factor Wiki} & 0   & 0.5855  \\
                     & 10  & 0.5895  \\
                     & 30  & 0.5858  \\
                     & 50  & 0.5879  \\
                     & 70  & 0.6058  \\
                     & 90  & 0.5873  \\
\midrule
\textbf{Factor News} & 0   & 0.5841 \\
                     & 10  & 0.5833  \\
                     & 30  & 0.5927  \\
                     & 50  & 0.5753  \\
                     & 70  & 0.6197 \\
                     & 90  & 0.5724  \\
\midrule
\textbf{TruthfulQA} & 0   & 23.62  41.21  19.33  \\
                    & 10  & 21.78  39.14  19.54  \\
                    & 30  & 21.54  46.67  24.19 \\
                    & 50  & 20.56  45.99  23.62  \\
                    & 70  & 25.21  53.70   26.50  \\
                    & 90  & 25.33  46.30   27.50 \\
\midrule
\textbf{HaluEval-Sum} & 0   & 18.94  26.06    \\
                    & 10  & 21.38  24.33    \\
                    & 30  & 27.15  27.25   \\
                    & 50  & 26.31  25.86    \\
                    & 70  & 22.41  23.12     \\
                    & 90  & 19.42  21.04    \\
\midrule
\textbf{HellaSwag} & 0   & 0.7801    \\
                   & 10  & 0.8140 \\
                   & 30  & 0.8424 \\
                   & 50  & 0.8372 \\
                   & 70  & 0.8239 \\
                   & 90  & 0.7945 \\
\midrule
\textbf{OpenbookQA} & 0   & 0.5141   \\
                    & 10  & 0.5123   \\
                    & 30  & 0.5325   \\
                    & 50  & 0.5567   \\
                    & 70  & 0.5581   \\
                    & 90  & 0.5423   \\
\midrule
\textbf{Race High} & 0   & 0.4320   \\
                   & 10  & 0.4379 \\
                   & 30  & 0.4388 \\
                   & 50  & 0.4545 \\
                   & 70  & 0.4637 \\
                   & 90  & 0.4614 \\
\midrule
\textbf{Race Middle} & 0   & 0.5740   \\
                     & 10  & 0.5731 \\
                     & 30  & 0.5989 \\
                     & 50  & 0.5926 \\
                     & 70  & 0.5843 \\
                     & 90  & 0.5933 \\
\bottomrule
\end{tabularx}
}
\end{table}

\begin{figure*}[t]
    \centering
    \includegraphics[width=1\linewidth]{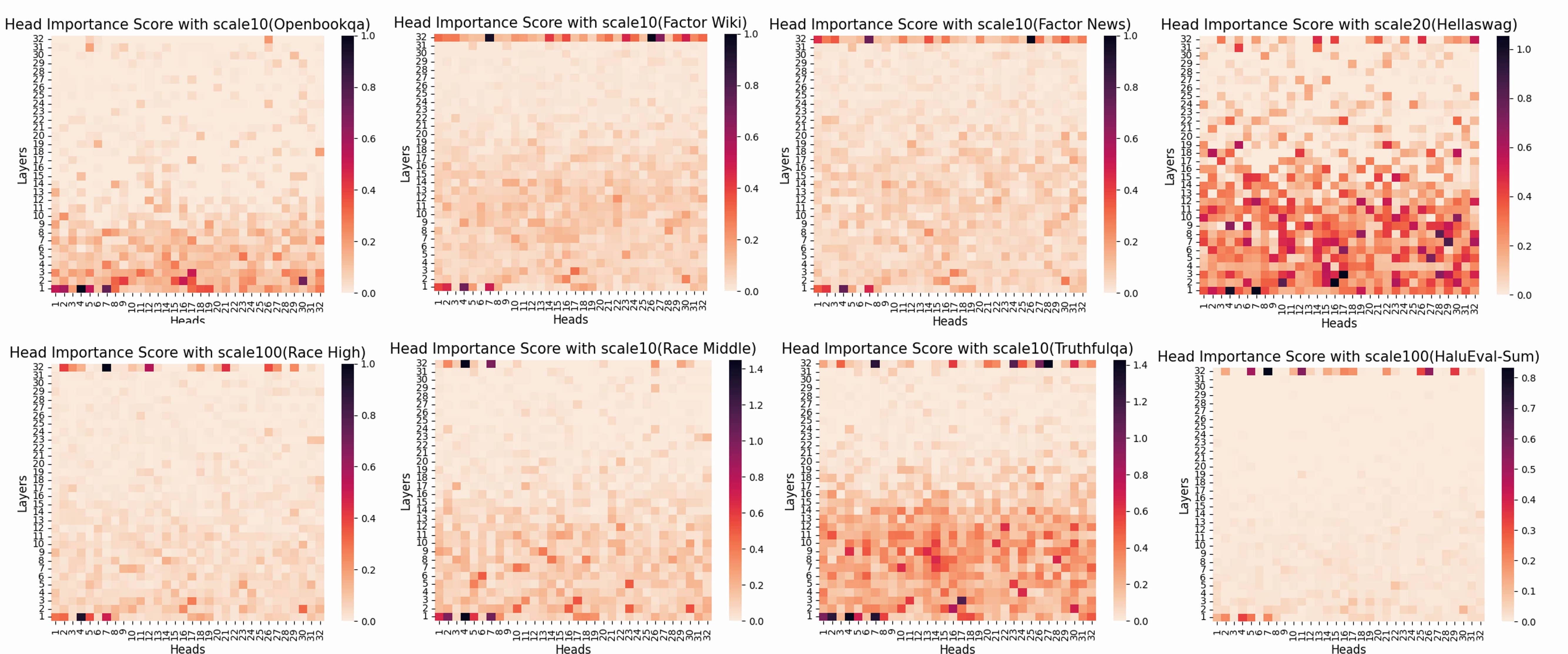}  
    \caption{Visualization of importance scores for attention heads,  used to select the inducing heads.}
    \label{fig:importance_scores}
    \vspace{-1em}
\end{figure*}

\subsection{Spearman Correlation Coefficient \(r\)}
\label{sec:spearman_correlation}

The Spearman correlation coefficient \(r\) is a non-parametric measure of statistical dependence between two variables. It assesses how well the relationship between two variables can be described using a monotonic function. In the context of our study, we use the Spearman correlation coefficient to quantify the relationship between the number of inducing heads and the performance of the downstream tasks. Specifically, we evaluate how the number of inducing heads affects the task performance. A higher correlation coefficient indicates that the number of inducing heads have a stronger impact on task performance.We calculate \(r\) across different tasks to observe how the number of inducing heads correlates with the performance metrics. The results are summarized in Figure \ref{fig:induced_heads_analysis}. The correlation values for each task are shown in Table \ref{tab:spearman_correlation}.

\section{ Inducing Head Analysis}
\label{sec: D appendix}

\subsection{Visualization of Importance Scores for Attention Heads}
\label{sec:visualization_importance_scores}

To identify the most relevant attention heads for inducing hallucinations, we visualize the importance scores for the attention heads, which are computed by combining the scores of right heads and wrong heads. These scores help us rank the heads from the most to the least important. Based on these rankings, we select the top-k heads to form the set of inducing heads.

The visualization of the importance scores, shown in Figure~\ref{fig:importance_scores}, illustrates the distribution of these scores across the heads. We use this scores to guide our selection of the top-k heads, where the most important heads are chosen for hallucination induction.

\subsection{Custom Metric for Inducing Head Selection}
\label{sec:custom_metric_induced_heads}

To evaluate the selection of inducing heads, we define a custom metric based on the overlap between the inducing heads and two key sets: the right heads and the wrong heads. Specifically, we aim to maximize the intersection between the inducing heads and the right heads while minimizing the intersection with the wrong heads.

The custom metric is computed as follows: for each set of inducing heads, we compute the overlap with the right and wrong heads sets and use these values to generate a score. Let \( H_r \) represent the set of right heads, \( H_w \) represent the set of wrong heads, and \( H_i \) represent the set of selected inducing heads. The custom metric score \( S_{\text{metric}} \) is computed as Equation \ref{mq:metric}.

\begin{equation}
S_{\text{metric}} = |H_i \cap H_r| - \beta \cdot |H_i \cap H_w|
\label{mq:metric}
\end{equation}

\begin{table}[H]
\centering
\small
\caption{Spearman correlation coefficient \(r\) for various tasks.}
\label{tab:spearman_correlation}
\begin{tabular}{lcc}
\toprule
\textbf{Task}        & \textbf{Spearman \(r\)} \\
\midrule
HellaSwag            & 0.2 \\
Race High            & 0.9429 \\
Race Middle          & 0.5429 \\
OpenBookQA           & 0.7714 \\
TruthfulQA (MC1)     & 0.4857 \\
TruthfulQA (MC2)     & 0.6571 \\
TruthfulQA (MC3)     & 0.8286 \\
Factor Wiki          & 0.4286 \\
Factor News          & 0.2571 \\
HaluEval-Sum(Acc-H)  & 0.3127 \\
HaluEval-Sum(Acc-A)  & 0.3512 \\
\bottomrule
\end{tabular}
\end{table}
Where:

- \( |H_i \cap H_r| \) is the number of inducing heads that overlap with the right heads

- \( |H_i \cap H_w| \) is the number of inducing heads that overlap with the wrong heads

- \( \beta \) is a hyperparameter that controls the penalty for overlap with the wrong heads

\begin{figure*}[t]
    \centering
    \includegraphics[width=1\linewidth]{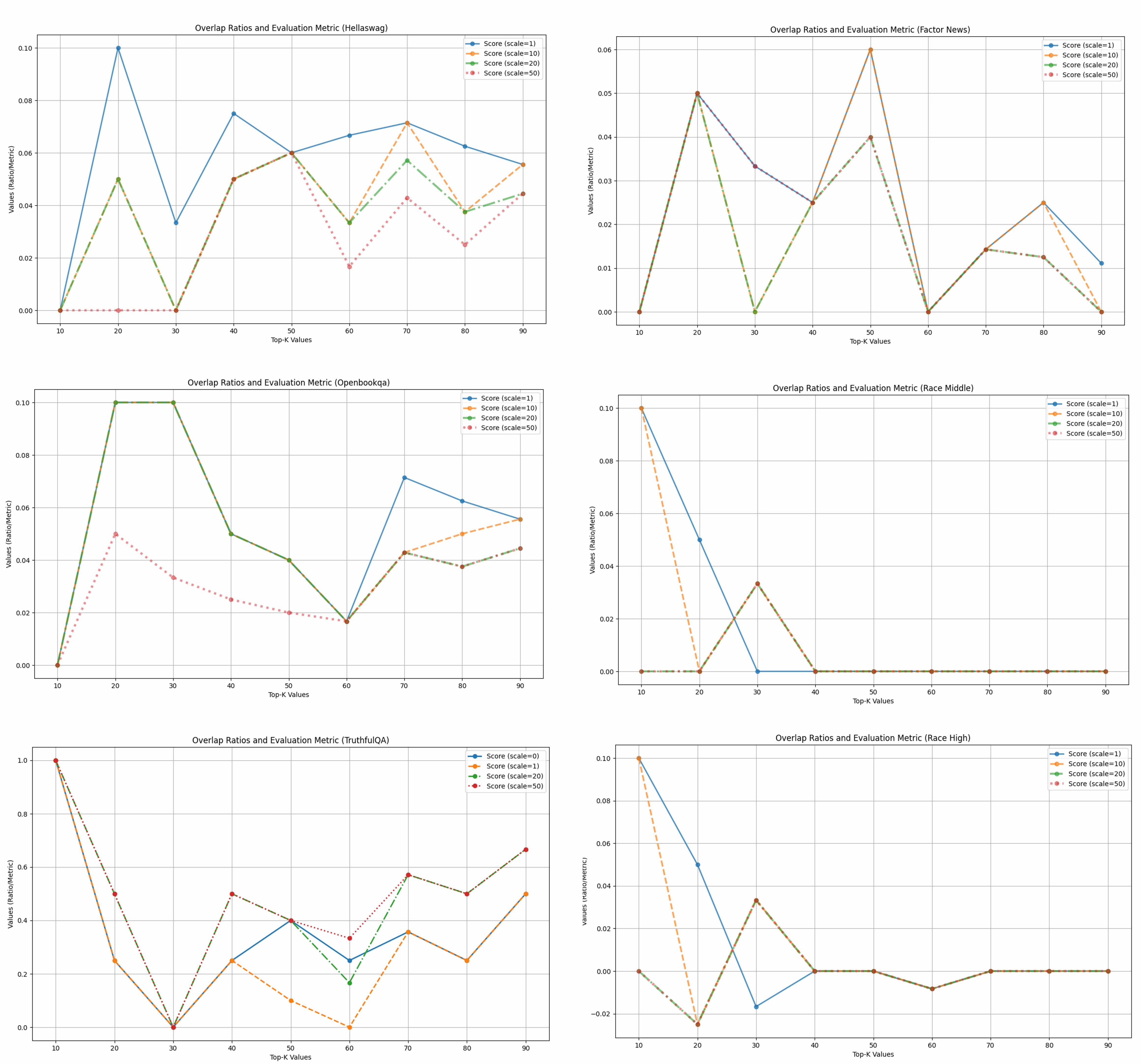}  
    \caption{Evaluation of inducing heads using our custom metric. The metric scores are plotted for various top-k and scale settings.}
    \label{fig:metric_results}
\end{figure*}

This score is maximized when the inducing heads align well with the right heads and avoid overlap with the wrong heads.
We evaluate this metric across different values of top-k and scale settings, and the results are shown in Figure~\ref{fig:metric_results}. This evaluation shows that the best configurations, as determined by our metric, align with the configurations yielding the best performance in our experiments.

\begin{table}[H]
\centering
\caption{Performance of Llama2-7b with Top-k inducing heads.}
\label{llama2_topk}
\small
\resizebox{\columnwidth}{!}{
\begin{tabularx}{\columnwidth}{lccccccccc}
\toprule
\textbf{Task} & \textbf{Top-k} & \textbf{Acc / MC} \\
\midrule
\textbf{Factor Wiki} & 0   & 0.5898 \\
                     & 10  & 0.5922 \\
                     & 30  & 0.5992 \\
                     & 50  & 0.6035 \\
                     & 70  & 0.6035 \\
                     & 90  & 0.6069 \\
\midrule
\textbf{Factor News} & 0   & 0.7200 \\
                     & 10  & 0.7249 \\
                     & 30  & 0.7249 \\
                     & 50  & 0.7288 \\
                     & 70  & 0.7346 \\
                     & 90  & 0.7307 \\
\midrule
\textbf{TruthfulQA}  & 0   & 28.51  43.30  22.40 \\
                     & 10  & 23.99  47.35  25.35 \\
                     & 30  & 21.78  47.13  23.76 \\
                     & 50  & 22.39  50.33  24.65 \\
                     & 70  & 23.99  51.28  25.89 \\
                     & 90  & 23.74  46.64  26.27 \\
\midrule
\textbf{HellaSwag}   & 0   & 0.7800 \\
                     & 10  & 0.8025 \\
                     & 30  & 0.8433 \\
                     & 50  & 0.8307 \\
                     & 70  & 0.8083 \\
                     & 90  & 0.8017 \\
\midrule
\textbf{OpenbookQA}  & 0   & 0.4841 \\
                     & 10  & 0.5012 \\
                     & 30  & 0.5181 \\
                     & 50  & 0.5021 \\
                     & 70  & 0.5223 \\
                     & 90  & 0.5124 \\
\midrule
\textbf{Race High}   & 0   & 0.4325 \\
                     & 10  & 0.4199 \\
                     & 30  & 0.4483 \\
                     & 50  & 0.4465 \\
                     & 70  & 0.4514 \\
                     & 90  & 0.4431 \\
\midrule
\textbf{Race Middle} & 0   & 0.5800 \\
                     & 10  & 0.5745 \\
                     & 30  & 0.5996 \\
                     & 50  & 0.6017 \\
                     & 70  & 0.5996 \\
                     & 90  & 0.5843 \\
\bottomrule
\end{tabularx}
}
\end{table}

\section{Comparison with Other Hallucination-Inducing Methods}
\label{sec: E appendix}

In this section, we compare HICD with several other hallucination-inducing methods. The goal of this comparison is to highlight the superior ability of HICD to induce "contrast-effective" hallucinations. The results are presented in Figure~\ref{fig:induced_heads_analysis} and Table~\ref{table2}.

\begin{itemize}
\setlength{\itemsep}{0pt}
\setlength{\parskip}{0pt}
\setlength{\topsep}{0pt}
\item \hypertarget{sec:prompt}{\textbf{Prompt-based}}: In line with the idea of inducing hallucinations, we leverage specially designed prompts to directly compel the model to generate fabricated information. We use the prompt: "\textit{You are a helpful, respectful but not honest assistant. You must generate false or fabricated information. This is very important to my career.}" This system prompt directs the model to intentionally produce false information, making it a useful tool for investigating the effects of hallucinations. By prompting the model in this manner, we can generate hallucinated outputs that are systematically different from the model’s original predictions, which allows us to perform contrastive analysis and study the impact of hallucinations on model performance.
\item \hypertarget{sec:sh2}{\textbf{SH2-based}}: Inspired by \cite{kai-etal-2024-sh2}, which selects tokens with high informational content and prepends them to the context. By repeating these high-information tokens, the model’s attention is shifted towards them, increasing their focus and improving the model's accuracy. In contrast, we are inspired by this idea, but we apply it in reverse. Instead of adding high-information tokens, we prepend low-information, low-relevance tokens to the context. This forces the model to shift its attention to these irrelevant tokens, which effectively induces hallucinations. Then we apply contrastive decoding to compare the hallucinated outputs with the original model outputs, thus mitigating hallucinations while preserving performance.

\item \hypertarget{sec:pasta}{\textbf{PASTA-based}}: Based on the Attention Steering method from \cite{zhang2023tell}, the PASTA-based method selects task-relevant attention heads and increases the attention weights of token positions corresponding to key context information. This technique improves the model’s attention to critical sentences or words, thus enhancing its contextual faithfulness. Following the ideas in PASTA, we manipulate the attention weights of low-information tokens, which have low relevance to the task or correctness of the output. By increasing the attention scores of low-relevance tokens, we intentionally shift the model's focus towards irrelevant or less informative words. This dispersion of attention results in the induction of hallucinations, as the model starts to generate content based on these non-essential tokens. We then apply contrastive decoding to compare the hallucinated outputs with the original model's outputs, effectively mitigating hallucinations while preserving overall model performance.

\item \hypertarget{sec:cut}{\textbf{Cut-based}}: The Cut-based method directly ignores the outputs of specific inducing heads by masking them, effectively forcing the model to disregard certain attention heads. This simple yet effective approach induces hallucinations by removing the influence of particular attention heads. After inducing hallucinations, contrastive decoding is applied to compare the hallucinated outputs with the original outputs.

\end{itemize}
As shown in Figure~\ref{fig:induced_heads_analysis}, in most tasks, the Cut-based method (blue lines) exhibits weaker performance in mitigating hallucinations at the optimal inducing head number compared to the Ave-based approach HICD (red lines). From the numerical results in Table~\ref{table2}, HICD consistently outperforms both the Prompt-based and the PASTA-based attention steering across all datasets. This is especially evident in tasks that require contextual faithfulness, where HICD achieves the best overall performance.

Although the SH2-based method for inducing hallucinations outperforms HICD on specific factuality tasks—such as the \textit{TruthfulQA} MC1 metric and the \textit{FACTOR} datasets—the overall results indicate that HICD has greater potential for inducing "contrast-effective" hallucinations. This advantage allows HICD to effectively mitigate hallucinations while maintaining superior performance across a wide range of evaluation tasks.

\section{In-domain vs Out-of-domain Inducing Head Evaluation}
\label{sec: F appendix}

We analyze the performance of out-of-domain inducing heads obtained from various datasets, with respect to the same task. As shown in Figure~\ref{fig:spearman}, the performance of these out-of-domain inducing heads varies depending on the correlation between the rankings of the inducing head scores. As the correlation between out-of-domain and in-domain inducing heads increases, the performance of the out-of-domain inducing heads becomes more similar to that of the in-domain inducing heads.

For example, the inducing heads from Race Middle, TruthfulQA, and HaluEval-Sum exhibit relatively high ranking similarity with OpenBookQA. This is evident from the data presented in Table~\ref{tab:metrics}, where the performance of the out-of-domain inducing heads from these datasets is closer to that of the in-domain OpenBookQA inducing heads compared to other out-of-domain heads. 

On the other hand, the Factor News tasks, exhibit a relatively lower Spearman correlation with other tasks. This results in a more uniform performance across the out-of-domain inducing heads from these datasets. This uniformity is accompanied by a notable gap in performance when compared to the in-domain inducing heads. In TruthfulQA task, out-of-domain heads from Factor News and Factor Wiki, which have lower correlations with in-domain heads, perform worse than other out-of-domain heads. At the same time, we observe that out-of-domain heads with a correlation greater than 50\% with in-domain heads exhibit a relatively larger performance improvement compared to those with a correlation below 50\%.

This analysis demonstrates that the inducing heads from out-of-domain datasets with higher correlation to the in-domain dataset yield more consistent with in-domain results.

\section{Norm Analysis and Token Confidence}
\label{sec: G appendix}

\subsection{Norm-Based Analysis}
\begin{figure*}[t]
    \centering
    \includegraphics[width=1\textwidth]{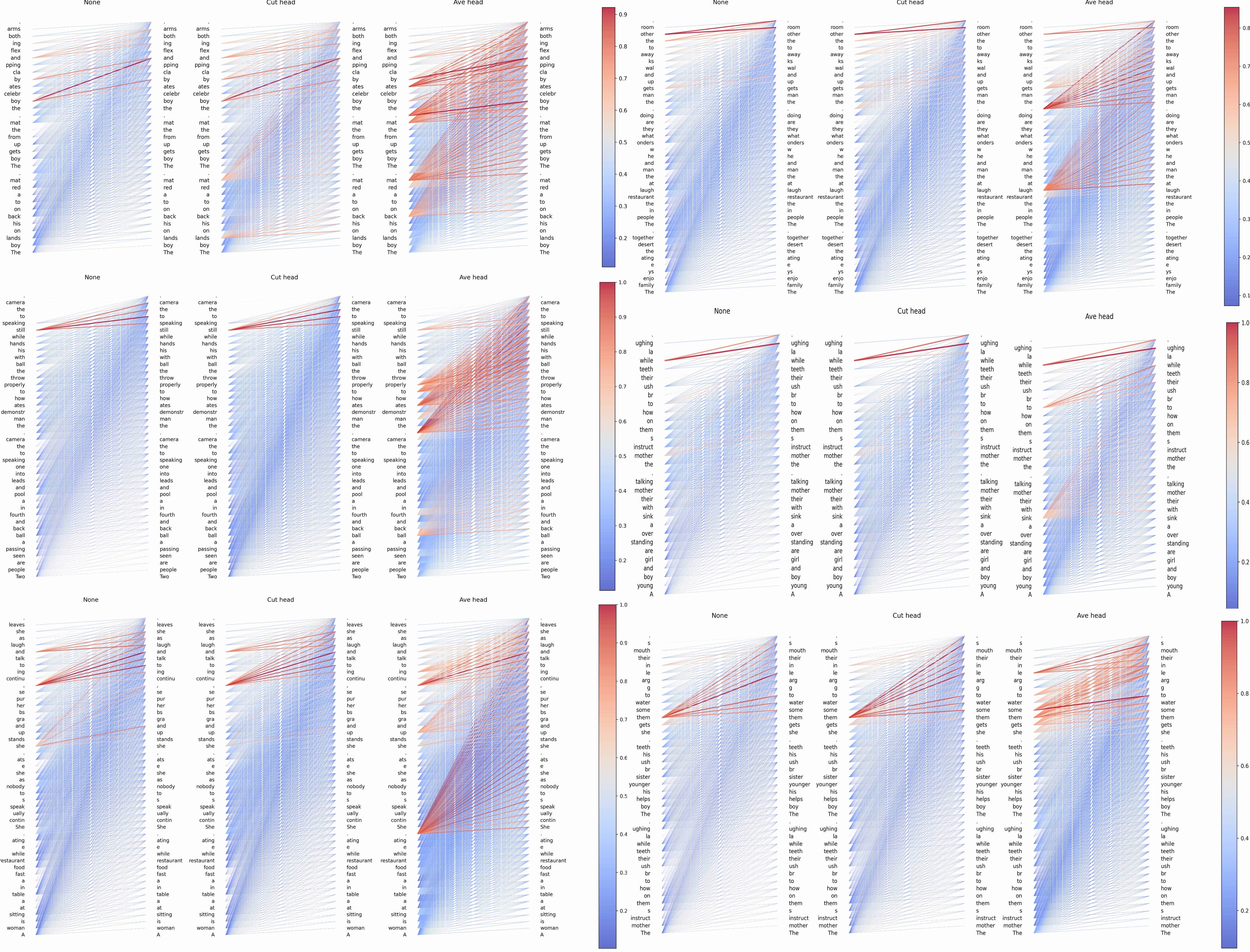}  
    \caption{Supplementary results showing the effect of different hallucination inducing methods on the information flow. This figure complements Figure~\ref{fig:saliency}, illustrating how Ave Head dispersion the attention distribution and enhances the effective of hallucinated outputs.}
    \label{fig:icl}
\end{figure*}
In Transformer models, the attention mechanism is essential for selecting relevant information from the input sequence. While attention weights \( \alpha \) are commonly used to measure the relevance of each token, recent work shows that the norm of the transformed input vectors, \( f(x) \), also plays a significant role in determining the final attention output. Specifically, the attention mechanism computes the output as a weighted sum of the transformed input vectors, where the transformed vector \( f(x_j) \) is calculated by applying a learned transformation to the input token \( x_j \), and the attention weight \( \alpha_{i,j} \) determines how much influence each token should have on the output:

\begin{equation}
y_i = \sum_{j=1}^{n} \alpha_{i,j} f(x_j)
\label{eq:example}
\end{equation}

In Equation \ref{eq:example}, \( f(x_j) \) represents the transformed vector of input token \( x_j \), and \( \alpha_{i,j} \) is the attention weight.

However, previous analyses based solely on attention weights \( \alpha \) overlook the critical role of \( f(x_j) \). As demonstrated in \cite{kobayashi-etal-2020-attention}, the attention weight-based analysis is insufficient because it does not account for the fact that the transformed vectors can have varying magnitudes, even if the attention weight is large.

To address this, the norm-based analysis that incorporates both the attention weights and the norms of the transformed vectors. Based on norm-based analysis, the model not only controls the contribution of different tokens through attention weights \( \alpha \) but also regulates the contribution levels of frequently occurring, low-information tokens by controlling the norm of \( f(x) \). In this framework, the final attention is not only governed by the attention weights \( \alpha \), but also by the magnitude of the transformed vectors \( f(x_j) \).

This norm-based perspective helps to better understand how Transformer models attend to different tokens, especially in cases where attention weights alone would lead to misleading interpretations. By adjusting the norms of these transformed vectors, we can change the influence of frequent, low-information tokens, leading to a more effective and nuanced attention allocation.
\begin{table*}[h!]
\centering
\caption{Examples of contextual prediction and their corresponding information flow in Figure~\ref{fig:icl}. The black portion of the text represents the context, and the blue portion shows the predicted tokens.}
\label{tab:saliency_examples}
\small  
\begin{tabular}{|p{0.45\textwidth}|p{0.45\textwidth}|}
\hline
\textbf{Context} & \textbf{Predicted Tokens} \\
\hline
The boy lands on his back on to a red mat. The boy gets up from the mat. the boy & \textcolor{blue}{celebrates by clapping and flexing both arms .} \\
\hline
A man is holding a pocket knife while sitting on some rocks in the wilderness. then he & \textcolor{blue}{takes a small stone from the flowing river and smashes it on another stone.} \\
\hline
Two people are seen passing a ball back and forth in a pool and leads into one speaking to the camera.the man & \textcolor{blue}{demonstrates how to properly throw the ball with his hands while still speaking to the camera.} \\
\hline
A woman is sitting at a table in a fast food restaurant while eating. She continually speaks to nobody as she eats. She& \textcolor{blue}{stands up and grabs her purse, continuing to talk and laugh as she leaves.} \\
\hline
The family enjoys eating the desert together. The people in the restaurant laugh at the man and he wonders what they are doing. the man & \textcolor{blue}{gets up and walks away to the other room.} \\
\hline
A young boy and girl are standing over a sink with their mother talking. the mother & \textcolor{blue}{instructs them on how to brush their teeth while laughing.} \\
\hline
The mother instructs them on how to brush their teeth while laughing. The boy helps his younger sister brush his teeth. she & \textcolor{blue}{gets them some water to gargle in their mouths.} \\
\hline
\end{tabular}
\end{table*}
\subsection{Token Confidence and Key Tokens}

In language models, the prediction of a token is typically driven by the context provided by previous tokens. The confidence of the model in its predictions can be quantified by the probability assigned to each token. We can define token confidence as the prediction probability of a token given its preceding context, \( p(\hat{x}_t) = p(\theta(\hat{x}_t | x_<t)) \), where \( \hat{x}_t \) is the token at position \( t \) and \( x_<t \) represents the context preceding it \cite{kai-etal-2024-sh2}.

Key tokens are defined as those that the model predicts with the lowest confidence. These tokens are the hardest for the model to predict and are considered to carry more semantic information, often representing critical content such as nouns, proper nouns, and adjectives. These tokens provide significant insight into the factual content of the text. The reasoning behind this is that tokens with lower confidence are harder for the model to infer, indicating that they are less predictable, and thus may contain more complex or factual information.

In contrast, high-confidence tokens, often function words such as prepositions or determiners, contribute less to the factual content of the sentence. They are generally easier for the model to predict, and their occurrence does not add much to the model's understanding of the facts.

Tokens with the highest informational content are those hardest to predict. The language model can benefit from giving more attention to these low-confidence tokens, as they are more likely to carry factual information, thus improving the factuality of the generated text.

\subsection{Saliency Matrix and Information Flow}
\label{sec:saliency_appendix}

We investigate the impact of attention map averaging on the model's outputs through the analysis of the saliency matrix \( I(i,j) \), where \( I(i,j) \) quantifies the importance of information flow from token \( i \) to token \( j \) \cite{wang-etal-2023-label}. The results reveals that Ave Head allows the model to generate hallucinations that appear more "plausible." 

Figure~\ref{fig:icl} provides further evidence supporting these observations. It shows how applying attention map averaging can alter the importance of information flow across tokens, thereby impacting the attention given to the tokens that need to be predicted based on the context, ultimately affecting the resulting outputs. The figure visualizes information flow, where the bottom row represents earlier tokens in the sentence and the top row represents later tokens. The connecting lines between tokens signify the strength of information flow, with thicker or more prominent lines indicating a stronger influence of one token on another.

Additionally, the examples provided in the figure are further detailed in Table~\ref{tab:saliency_examples}, which lists the context and the corresponding predicted tokens.

\section{Additional Results on Other Models}
\label{sec: G appendix}

We conducted experiments on HICD and the relevant baselines using \textbf{Qwen-7B}, \textbf{Mistral-7B-v0.3}, and \textbf{LLaMA-3-8B-Instruct}. The evaluation results on \textbf{faithfulness-related task datasets} are presented in Table~\ref{tab:faithfulness}. HICD achieves the best overall performance on faithfulness tasks compared to other methods. Therefore, the proposed HICD method, based on efficient contrastive hallucination induction, effectively mitigates contextual faithfulness hallucinations across multiple models.

\begin{table}[htbp]
\centering
\caption{Performance on Tasks across Models}
\label{tab:faithfulness}
\scriptsize 
\renewcommand{\arraystretch}{1.1}
\setlength{\tabcolsep}{3pt} 
\begin{tabular}{lccc}
\toprule
\textbf{Model} & \textbf{Hellaswag} & \textbf{Race (M/H)} & \textbf{OpenbookQA} \\
\midrule
Qwen-7B            & 0.787 & 0.561 / 0.447 & 0.492 \\
\quad + CAD        & --    & 0.571 / 0.462 & 0.522 \\
\quad + DoLA       & 0.765 & 0.566 / 0.445 & 0.503 \\
\quad + HICD       & \textbf{0.801} & \textbf{0.573 / 0.471} & \textbf{0.534} \\
\midrule
Mistral-7B-v0.3    & 0.857 & 0.677 / 0.541 & 0.602 \\
\quad + CAD        & --    & 0.667 / 0.561 & 0.632 \\
\quad + DoLA       & 0.855 & 0.671 / 0.543 & 0.616 \\
\quad + HICD       & \textbf{0.871} & \textbf{0.689 / 0.557} & \textbf{0.634} \\
\midrule
LLaMA3-8B-Instruct & 0.817 & 0.671 / 0.518 & 0.538 \\
\quad + CAD        & --    & 0.702 / 0.543 & 0.562 \\
\quad + DoLA       & 0.822 & 0.685 / 0.527 & 0.564 \\
\quad + HICD       & \textbf{0.864} & \textbf{0.669 / 0.549} & \textbf{0.581} \\
\bottomrule
\end{tabular}
\end{table}

Although HICD is primarily designed to mitigate faithfulness hallucinations, we also report its performance on \textbf{factual consistency tasks} across different models. As shown in Table~\ref{tab:factual}, HICD demonstrates improvements in factual consistency across various models.

\begin{table}[htbp]
\centering
\caption{Performance on Factual Consistency Tasks}
\label{tab:factual}
\scriptsize 
\renewcommand{\arraystretch}{1.1}
\setlength{\tabcolsep}{2pt}
\begin{tabular}{lcc}
\toprule
\textbf{Model} & \textbf{TruthfulQA (MC1/2/3)} & \textbf{Factor (WIKI / NEWS)} \\
\midrule
Qwen-7B          & 30.59 / 46.95 / 25.06 & 58.78 / 69.88 \\
\quad + HICD     & \textbf{25.94} / \textbf{47.30} / \textbf{27.29} & \textbf{60.21} / \textbf{70.15} \\
\midrule
Mistral-7B-v0.3  & 49.71 / 63.23 / 37.15 & 59.31 / 74.22 \\
\quad + HICD     & \textbf{49.75} / \textbf{65.63} / \textbf{37.92} & \textbf{61.80} / \textbf{76.58} \\
\midrule
LLaMA-3-8B-Instruct & 39.31 / 56.91 / 30.43 & 61.02 / 74.51 \\
\quad + HICD     & \textbf{40.21} / \textbf{59.79} / \textbf{34.45} & \textbf{64.43} / \textbf{75.50} \\
\bottomrule
\end{tabular}
\end{table}

\textbf{Comparison with other baseline}  
We also evaluate the performance of two additional baselines ITI\cite{NEURIPS2023_81b83900} and DeCoRe\cite{gema2024decore} on multiple datasets using LLaMA-3-8B-Instruct and Mistral-7B-v0.3. The results are presented in Table~\ref{tab:iti_decore}.

\begin{table}[htbp]
\centering
\caption{Comparison with ITI and DeCoRe }
\label{tab:iti_decore}
\scriptsize
\renewcommand{\arraystretch}{1.1}
\setlength{\tabcolsep}{2pt}
\begin{tabular}{lcc}
\toprule
\textbf{Model+Method} & \textbf{TruthfulQA (MC1/2/3)} & \textbf{Factor (WIKI/NEWS)} \\
\midrule
Llama3-8B-Instruct & 39.31 / 56.91 / 30.43 & 61.02 / 74.51 \\
+ ITI & \textbf{41.37} / 59.78 / \textbf{34.81} & 61.06 / 72.59 \\
+ DeCoRe & 38.43 / 55.86 / 30.31 & 62.33 / 75.45 \\
+ HICD & 40.21 / \textbf{59.79} / 34.45 & \textbf{64.43} / \textbf{75.50} \\
\midrule
Mistral-7b-v0.3 & 49.71 / 63.23 / 37.15 & 59.31 / 74.22 \\
+ ITI & \textbf{51.23} / \textbf{65.78} / \textbf{39.32} & 60.85 / 73.34 \\
+ DeCoRe & 48.23 / 59.14 / 35.21 & 61.32 / \textbf{76.70} \\
+ HICD & 49.75 / 65.63 / 37.92 & \textbf{61.82} / 76.51 \\
\bottomrule
\end{tabular}
\end{table}

The results indicate that \textbf{DeCoRe}, which generates contrastive outputs by retrieving and attending to external content via retrieval heads, shows unstable performance across datasets. While it achieves minor improvements on some tasks, it performs worse than vanilla outputs in others. This may stem from its reliance on pre-trained weights without sufficient task adaptation.

\textbf{ITI}, although strong on TruthfulQA, underperforms on most other datasets. We attribute this to its fine-tuning on the TruthfulQA dataset, which likely overfits it to that specific factuality task, impairing generalization.

In contrast, \textbf{HICD} employs a task-driven hallucination induction mechanism via attention dispersion and inducing head selection. This approach not only delivers more consistent improvements across tasks but also ensures better generalization capability. HICD proves to be more broadly applicable and effective than ITI and DeCoRe in mitigating hallucinations and improving overall output faithfulness.

\section{Computational Cost and Efficiency Analysis}
\label{sec:computation_appendix}

To better illustrate the superiority of HICD, we present the trade-off between performance improvement and computational cost. Table~\ref{tab:efficiency} shows the computational cost metrics for various methods, alongside the average performance metrics across all tasks.

\begin{table}[htbp]
\centering
\caption{Comparison of Efficiency and Performance}
\label{tab:efficiency}
\scriptsize
\renewcommand{\arraystretch}{1.1}
\setlength{\tabcolsep}{3pt}
\begin{tabular}{lcccc}
\toprule
\textbf{Method} & \textbf{Latency (ms) ↓} & \textbf{Throughput ↑} & \textbf{TFLOPS ↓} & \textbf{Avg. Metric ↑} \\
\midrule
LLaMA-7B   & 420 & 152 & 4.21 & 0.4317 \\
DoLA       & 435 & 147 & 4.76 & 0.4614 \\
CAD        & 690 & 92  & 8.38 & 0.4769 \\
HICD       & 613 & 104 & 6.07 & 0.4979 \\
\bottomrule
\end{tabular}
\end{table}

HICD demonstrates its superiority through the highest average metric (0.4979), outperforming all baseline methods. While its inference latency (613 ms) is slightly higher than that of DoLA (435 ms), it maintains competitive throughput (104 requests/s) and lower TFLOPS than CAD, reflecting a strong trade-off between efficiency and performance. This makes HICD the most effective and balanced method overall.

\end{document}